\documentclass{article}

\usepackage{microtype}
\usepackage{graphicx}
\usepackage{booktabs} 

\usepackage{hyperref}

\usepackage{amssymb}

\usepackage{amsmath}
\usepackage{amsfonts}
\usepackage{algorithm}
\usepackage{algorithmic}

\newtheorem{theorem}{Theorem}
\newtheorem{lemma}{Lemma}

\usepackage{times}
\usepackage{epsfig}
\usepackage{graphicx}
\usepackage{amsmath}
\usepackage{amssymb}
\usepackage{subfig}

\usepackage{amsmath}
\usepackage{amsfonts}
\usepackage{algorithm}
\usepackage{algorithmic}
\usepackage{color}

\usepackage{lineno,hyperref}
\usepackage{amsfonts}
\usepackage{multirow}
\usepackage{bm}

\usepackage{diagbox}
\usepackage{multirow}

\usepackage{times}
\usepackage{soul}
\usepackage{url}
\usepackage{algorithm}
\usepackage{algorithmic}
\usepackage{subfig}
\usepackage{lineno,hyperref}
\usepackage{amsfonts}
\usepackage{multirow}
\usepackage{bm}
\modulolinenumbers[5]
\usepackage{rotating}
\usepackage{booktabs}
\usepackage{makecell}

\newtheorem{corollary}{Corollary}
\newtheorem{assumption}{Assumption}

\usepackage[accepted]{icml2021}

\usepackage{lipsum}
\usepackage{todonotes}

\newcommand{\reza}[1]{\textcolor{black}{ {#1}}}

\newcommand{\miao}[1]{\textcolor{black}{ {#1}}}

\icmltitlerunning{iDARTS: Differentiable Architecture Search with Stochastic Implicit Gradients}

\begin{document}

\marginparwidth=20mm

\twocolumn[
\icmltitle{iDARTS: Differentiable Architecture Search with Stochastic Implicit Gradients}




\begin{icmlauthorlist}
\icmlauthor{Miao Zhang}{mo,uts}
\icmlauthor{Steven Su}{uts}
\icmlauthor{Shirui Pan}{mo}
\icmlauthor{Xiaojun Chang}{mo}
\icmlauthor{Ehsan Abbasnejad}{ad}
\icmlauthor{Reza Haffari}{mo}
\end{icmlauthorlist}

\icmlaffiliation{mo}{Faculty of Information Technology, Monash University, Australia}
\icmlaffiliation{uts}{Faculty of Engineering and Information Technology, University of Technology Sydney, Australia}
\icmlaffiliation{ad}{Australian Institute for Machine Learning, University of Adelaide, Australia}

\icmlcorrespondingauthor{Shirui Pan}{Shirui.Pan@monash.edu}

\icmlkeywords{Machine Learning, ICML}

\vskip 0.3in
]

\newcommand{\pan}[1]{\textcolor{orange}{\small\textbf{[SP]} #1$\triangleleft$}}



\printAffiliationsAndNotice{}  

\begin{abstract}

\textit{Differentiable ARchiTecture Search} (DARTS) has recently become the mainstream of neural architecture search (NAS) due to its efficiency and simplicity. 
With a gradient-based bi-level optimization, DARTS alternately optimizes the inner model weights and the outer architecture parameter in a weight-sharing supernet. 
%
A key challenge to the scalability and quality of the learned architectures is the need for differentiating through the inner-loop optimisation. 
While much has been discussed about several potentially fatal factors in DARTS, 
the architecture gradient, a.k.a. hypergradient, has received less attention.
%
%
In this paper, we tackle the hypergradient computation in DARTS based on the implicit function theorem, making it only depends on the obtained solution to the inner-loop optimization and agnostic to the optimization path. 
%
To further reduce the computational requirements, we formulate a stochastic hypergradient approximation for differentiable NAS, and theoretically show that the architecture optimization with the proposed method, named iDARTS, is expected to converge to a stationary point. 
Comprehensive experiments on two NAS benchmark search spaces and the common NAS search space verify the effectiveness of our proposed method. It leads to architectures outperforming, with large margins, those learned by the baseline methods. 
\end{abstract}


\section{Introduction}
Neural Architecture Search (NAS) is an efficient and effective method on automating the process of neural network design, with achieving remarkable success on image recognition \cite{tan2019efficientnet,LiWWLLC21,LiPYWLLC20}, language modeling \cite{jiang2019improved}, and other deep learning applications \cite{ren2020comprehensive,cheng2020hierarchical,chen2019detnas,HuZCL21,ZhuLZCL21,RenXCXLC21}. The early NAS frameworks are devised via reinforcement learning (RL) \cite{pham2018efficient} or evolutionary algorithm (EA) \cite{real2018regularized} to directly search on the discrete space. To further improve the efficiency, a recently proposed \textit{Differentiable ARchiTecture Search} (DARTS) \cite{liu2018darts} adopts the continuous relaxation to convert the operation selection problem into the continuous magnitude optimization for a set of candidate operations. By enabling the gradient descent for the architecture optimization, DARTS significantly reduces the search cost to several GPU hours \cite{liu2018darts,xu2019pcdarts,GDAS}.

Despite its efficiency, more current works observe that DARTS is somewhat unreliable \cite{zela2019understanding,chen2020stabilizing,li2019random,sciuto2019evaluating,zhang2020one,LiTWPWLC21,zhang2020overcoming} since it does not consistently yield excellent solutions, performing even worse than random search in some cases.  \citet{zela2019understanding} attribute the failure of DARTS to its supernet training, with empirically observing that the instability of DARTS is highly correlated to the dominant eigenvalue of the Hessian matrix of the validation loss with respect to architecture parameters. 
On the other hand, \citet{Rethinking2021} turn to the magnitude-based architecture selection process, who empirically and theoretically show the magnitude of architecture parameters does not necessarily indicate how much the operation contributes to the supernet’s performance.  \citet{chen2020stabilizing} observe a precipitous validation loss landscape with respect to architecture parameters, which leads to a dramatic performance drop when discretizing the final architecture for the operation selection. Accordingly, they propose a perturbation based regularization to smooth the loss landscape and improve the stability.

While there are many variants on improving the DARTS from various aspects, limited research attention has been paid to the approximation of the architecture parameter gradient, which is also called the outer-loop gradient or hypergradient. To fill the gap, this paper focuses on the hypergradient calculation in the differentiable NAS. The main contribution of this work is the development of the differentiable architecture search with stochastic implicit gradients (iDARTS). Specifically, we first revisit the DARTS from the bi-level optimization perspective and utilize the implicit function theorem (IFT) \cite{bengio2000gradient,lorraine2020optimizing}, instead of the one-step unroll learning paradigm adopted by DARTS, to calculate the architecture parameter gradient.  This IFT based hypergradient depends only on the obtained solution to the inner-loop optimization weights rather than the path taken, thus making the proposed method memory efficient and practical with numerous inner optimization steps. Then, to avoid calculating the inverse of the Hessian matrix with respect to the model weights, we utilize the Neumann series \cite{lorraine2020optimizing} to approximate this inverse and propose an approximated hypergradient for DARTS accordingly.
After that, we devise a stochastic approximated hypergradient to relieve the computational burden further, making the proposed method applicable to the differentiable NAS. We theoretically demonstrate that, under some mild assumptions \cite{ghadimi2018approximation,couellan2016convergence,grazzi2020convergence} on the inner and outer loss functions, the proposed method is expected to converge to a stationary point with small enough learning rates. Finally, we verify the effectiveness of the proposed approach on two NAS benchmark datasets and the common DARTS search space.

We make the following contributions:

\begin{itemize}
\item This paper deepens our understanding of the hypergradient calculation in the differentiable NAS. We reformulated the hypergradient in the differentiable NAS with the implicit function theorem (IFT), which can thus gracefully handle many inner optimization steps without increasing the memory requirement.

\item To relieve the heavy computational burdens, we consider a Neumann-approximation for the IFT based differentiable NAS. Further, to  make the implicit hypergradient practical for differentiable NAS, we formulate a stochastic hypergradient with the Neumann-approximation.
\item We provide a theoretical analysis of the proposed method and demonstrate that the proposed method is expected to converge to a stationary point when applied to differentiable NAS. Extensive experiments verify the effectiveness of the proposed method which significantly improves the performance of the differentiable NAS baseline on the NAS-Bench-1Shot1 and the NAS-Bench-201 benchmark datasets and the common DARTS search space.
\end{itemize}

\section{Preliminaries: DARTS and Bi-level Optimization}
\label{sec2}

Existing differentiable NAS methods mostly leverage the weight sharing and continuous relaxation to enable the gradient descent for the discrete architecture search, significantly improving the search efficiency. DARTS \cite{liu2018darts} is one of the most representative differentiable NAS methods, which utilizes the continuous relaxation to convert the discrete operation selection into the magnitude optimization for a set of candidate operations. Typically, NAS searches for cells to stack the full architecture, where a cell structure is represented as a directed acyclic graph (DAG) with $\textit{N}$ nodes. NAS aims to determine the operations and corresponding connections for each node, while DARTS applies a \textbf{softmax} function to calculate the magnitude of each operation, transforming the operation selection into a continuous magnitude optimization problem:
\begin{equation}\nonumber
\resizebox{.85\linewidth}{!}{$
    \displaystyle
\mathbf{X}_n=\sum_{0\leq s<n}\sum_{o=1}^{\left | \mathcal{O} \right |}\bar{\alpha}^{(s,n)}_oo(\mathbf{X}_s),\qquad \bar{\alpha}^{(s,n)}_o=\frac{\textup{exp}(\alpha_{o}^{s,n})}{\sum_{o'\in \mathcal{O}}\textup{exp}(\alpha_{o'}^{s,n})},
$}
\end{equation}
where $\mathbf{X}_n$ is the output of node $n$, $\mathcal{O}$ contains all candidate operations, and the output of each node is the weighted sum of its previous nodes' outputs affiliated with all possible operations. In this way, DARTS transforms the discrete architecture search into optimizing the continuous magnitude $\hat{\alpha}_{o}^{s,n}$, enabling gradient descent for the architecture optimization. A discrete architecture is obtained by applying an \textbf{argmax} function to the magnitude matrix after the differentiable architecture optimization.

The optimization in DARTS is based on the bi-level optimization formulation \cite{colson2007overview,liu2018darts}:
\begin{equation} \label{eq:darts_1}
 \begin{aligned}
&\underset{\alpha}{\textup{min}}\ \ \ \mathcal{L}_{val}(w^*(\alpha),\alpha) \\
&\textup{s.t.}\ \ \ w^*(\alpha)=\textup{argmin}_w\ \mathcal{L}_{train}(w,\alpha),
\end{aligned}
\end{equation}
where $\alpha$ is the continuous architecture representation and $w$ is the supernet weights. We indicate the $\mathcal{L}_{val}$ as $\mathcal{L}_{2}$ and the $\mathcal{L}_{train}$ as $\mathcal{L}_{1}$ in the remaining text for convenience. The nested formulation in DARTS is the same as the gradient-based hyperparameter optimization with bi-level optimization \cite{franceschi2018bilevel,maclaurin2015gradient,pedregosa2016hyperparameter}, where the inner-loop is to train the network parameter $w$ and the outer-loop is to optimize the architecture parameter $\alpha$. The gradient of the outer-loop for DARTS is then calculated as:
\begin{equation} \label{eq:bileve_hypergradient}
\nabla_{\alpha}\mathcal{L}_2=(\frac{\partial \mathcal{L}_2}{\partial \alpha}+\frac{\partial \mathcal{L}_2}{\partial w}\frac{\partial w^*(\alpha)}{\partial \alpha}).
\end{equation}
{DARTS considers the one-step unroll learning paradigm \cite{liu2018darts,rajeswaran2019meta} for the hypergradient calculation. This is done by taking a single step  in optimising $w$ instead of the optimal $w^*$.}

Different from the majority of existing works that attributes the failure of DARTS to its supernet optimization 
\cite{zela2019understanding,benyahia2019overcoming}, or the final discretization with \textbf{argmax} \cite{chen2020stabilizing,ruochen2021}, this paper revisits DARTS from the perspective of the hypergradient calculation $\nabla_{\alpha}\mathcal{L}_2$. Rather than considering the one-step unroll learning paradigm \cite{liu2018darts,finn2017model}, this paper utilizes the implicit function theorem (IFT) \cite{bengio2000gradient,lorraine2020optimizing} to reformulate the hypergradient calculation in DARTS. In the following subsection, we first recap the hypergradient calculation with different paradigms for DARTS. 




\section{Hypergradient: From Unrolling to iDARTS}



\textbf{One-step unrolled differentiation.} The one-step unroll learning paradigm, as adopted by DARTS, is commonly used in the bi-level optimization based applications, including meta-learning \cite{finn2017model}, hyperparameter optimization \cite{luketina2016scalable}, generative adversarial networks \cite{metz2016unrolled}, and neural architecture search \cite{liu2018darts}, as it simplifies the hypergradient calculation and makes the bi-level optimization formulation practical for large-scale network learning. \miao{As described in the Section \ref{sec2}, the one-step unroll learning paradigm restricts the inner-loop optimization with only one step training. Differentiating through the inner learning procedure with one step $w^*(\alpha)=w-\gamma \nabla_w \mathcal{L}_1$, and obtaining $\frac{\partial w^*(\alpha)}{\partial \alpha}=-\gamma \frac{\partial^2 \mathcal{L}_1}{\partial \alpha \partial w}$, DARTS calculates the hypergradient as:
\begin{equation} \label{eq:darts_hypergradient}
\nabla_{\alpha}\mathcal{L}_2^{DARTS}=\frac{\partial \mathcal{L}_2}{\partial \alpha}-\gamma \frac{\partial \mathcal{L}_2}{\partial w}\frac{\partial^2 \mathcal{L}_1}{\partial \alpha \partial w},
\end{equation}
where $\gamma$ is the inner-loop learning rate for $w$.}


\textbf{Reverse-mode back-propagation.} Another direction of computing hypergradient is the reverse-mode \cite{franceschi2017forward,shaban2019truncated}, which trains the inner-loop with enough steps to reach the optimal points for the inner optimization. This paradigm assumes $T$-step is large enough to adapt $w(\alpha)$ to $w^*(\alpha)$ in the inner-loop. Defining $\Phi$ as a step of inner optimization that $w_t(\alpha)=\Phi(w_{t-1},\alpha)$, and defining $Z_t=\nabla_{\alpha} w_t(\alpha)$, we have:
\begin{equation} \nonumber
Z_t=A_t Z_{t-1}+B_t,
\end{equation}
where $A_t=\frac{\partial \Phi(w_{t-1},\alpha)}{\partial w_{t-1}}$, and $B_t=\frac{\partial \Phi(w_{t-1},\alpha)}{\partial \alpha}$.

\miao{Then the hypergradient of DARTS with the reverse model could be formulated as:
\begin{equation} \label{eq:all_step}
\begin{aligned}
&\nabla_{\alpha}\mathcal{L}_2^{Reverse}=\frac{\partial \mathcal{L}_2}{\partial \alpha}+\frac{\partial \mathcal{L}_2}{\partial w_T}(\sum_{t=0}^{T}B_{t}A_{t+1}...A_T).\\
\end{aligned}
\end{equation}}



Although the reverse-mode bi-level optimization is easy to implement, the memory requirement linearly increases with the number of steps $T$ \cite{franceschi2017forward} as it needs to store all intermediate gradients, making it impractical for deep networks. Rather than storing the gradients for all steps, a recent work \cite{shaban2019truncated} only uses the last $K$-step ($K<<T$) gradients to approximate the exact hypergradient, which is called the truncated back-propagation. Based on the $K$-step truncated back-propagation, the hypergradient for DARTS could be described as:
\begin{equation} \label{eq:reverse_k_step}
\resizebox{.85\linewidth}{!}{$
    \displaystyle
h_{T-K}=\frac{\partial \mathcal{L}_2}{\partial \alpha}+\frac{\partial \mathcal{L}_2}{\partial w_T}Z_T=\frac{\partial \mathcal{L}_2}{\partial \alpha}+\frac{\partial \mathcal{L}_2}{\partial w_T}(\sum_{t=T-K+1}^{T}B_{t}A_{t+1}...A_T).\\
$}
\end{equation}


The lemmas in \cite{shaban2019truncated} show that $h_{T-K}$ is a sufficient descent direction for the outer-loop optimization. 
\begin{lemma}
\label{lemma_ksep}
\cite{shaban2019truncated}. For all $K\geq 1$, with $T$ large enough and $\gamma$ small enough, $h_{T-K}$ is a sufficient descent direction that, i.e. $h_{T-K}^\top \nabla_{\alpha}\mathcal{L}_2 \geq \Omega (\left \|  \nabla_{\alpha}\mathcal{L}_2 \right \|^2)$.
\end{lemma}

\textbf{iDARTS: Implicit gradients differentiation.}
Although $h_{T-K}$ significantly decreases memory requirements, it still needs to store $K$-step gradients, making it impractical for differentiable NAS. In contrast, by utilizing the implicit function theorem (IFT), the hypergradient can be calculate without storing the intermediate gradients \cite{bengio2000gradient,lorraine2020optimizing}. The IFT based hypergradient for DARTS could be formulated as the following lemma.
\begin{lemma}
\label{lemma_ift}
Implicit Function Theorem: Consider $\mathcal{L}_1$, $\mathcal{L}_2$, and $w^*(\alpha)$ as defined in Eq.\eqref{eq:darts_1}, and with $\frac{\partial\mathcal{L}_{1}(w^*,\alpha)}{\partial w}=0$, we have 
\begin{equation} \label{eq:ift_hypergradient}
\nabla_{\alpha}\mathcal{L}_2=\frac{\partial \mathcal{L}_2}{\partial \alpha}-\frac{\partial \mathcal{L}_2}{\partial w}\left [ \frac{\partial^2 \mathcal{L}_1}{\partial w\partial w} \right ]^{-1}\frac{\partial^2 \mathcal{L}_1}{\partial \alpha \partial w}.
\end{equation}
\end{lemma}

This is also called as implicit differentiation theorem \cite{lorraine2020optimizing}. However, for a large neural network, it is hard to calculate the inverse of Hessian matrix in Eq.\eqref{eq:ift_hypergradient}, and one common direction is to approximate this inverse. Compared with Eq.\eqref{eq:ift_hypergradient}, the hypergradient of DARTS \cite{liu2018darts} in Eq.\eqref{eq:darts_hypergradient}, which adopts the one-step unrolled differentiation, simply uses an identity to approximate the inverse $\left [ \frac{\partial^2 \mathcal{L}_1}{\partial w\partial w} \right ]^{-1}=\gamma I$. This naive approximation is also adopted by \cite{luketina2016scalable,balaji2018metareg,nichol2018first}. 
In contrast, \cite{rajeswaran2019meta,pedregosa2016hyperparameter} utilize the conjugate gradient (CG) to convert the approximation of the inverse to solving a linear system with $\delta$-optimal solution, with applications to the hyperparameter optimization and meta-learning. 

Recently, the Neumann series is introduced to approximate the inverse in the hyperparameter optimization \cite{lorraine2020optimizing} for modern and deep neural networks since it is a more stable alternative to CG and useful in stochastic settings. This paper thus adopts the Neumann series for the inverse approximation and proposes an approximated hypergradient for DARTS accordingly. A stochastic approximated hypergradient is further devised to fit with differentiable NAS and relieve the computational burden, which is called Differentiable Architecture Search with Stochastic Implicit Gradients (\textbf{iDARTS}). We theoretically show the proposed method converges in expectation to a stationary point for the differentiable architecture search with small enough learning rates. A detailed description and analysis of \textbf{iDARTS} follow in the next section.

\section{Stochastic Approximations in iDARTS}
\label{sec3}

As described, our \textbf{iDARTS} utilizes the Neumann series to approximate the inverse of the Hessian matrix for the hypergradient calculation in the IFT-based bi-level optimization of NAS. 
We further consider a stochastic setting where  the Neumann approximation is computed based on minibatch samples, instead of the full dataset, enabling scalability to large datasets, similar to standard-practice in deep learning.

This section starts by analyzing the bound of the proposed hypergradient approximation, and then shows the convergence property of the proposed stochastic approximated hypergradient for differentiable NAS. 

Before our analysis, we give the following common assumptions in the bi-level optimization.\footnote{Similar assumptions are also considered in \cite{couellan2016convergence,ghadimi2018approximation,grazzi2020convergence,grazzi2020iteration}.}

\begin{assumption}
\label{assumption1}
For the outer-loop function $\mathcal{L}_2$:
\begin{enumerate}
  \item For any  $w$ and $\alpha$, $\mathcal{L}_2(w,\cdot)$ and $\mathcal{L}_2(\cdot ,\alpha)$ are bounded below.
  \item For any $w$ and $\alpha$, $\mathcal{L}_2(w,\cdot)$ and $\mathcal{L}_2(\cdot ,\alpha)$ are Lipschitz continuous with constants $L^{w}_2> 0$ and  $L^{\alpha}_2> 0$.  
  \item For any $w$ and $\alpha$, $\nabla_{w}\mathcal{L}_2(w,\cdot)$ and $\nabla_{\alpha}\mathcal{L}_2(\cdot ,\alpha)$ are Lipschitz continuous with constants $L^{\nabla_w}_2> 0$ and $L^{\nabla_\alpha}_2> 0$ with respect to $w$ and $\alpha$.

\end{enumerate}
\end{assumption}

\begin{assumption}
\label{assumption2}
For the inner-loop function $\mathcal{L}_1$
\begin{enumerate}
  \item  $\nabla_{w}\mathcal{L}_1$ is Lipschitz continuous with respect to $w$ with constant $L_1^{\nabla_w}> 0$.  
  \item  The function $w:\alpha \rightarrow w(\alpha)$ is Lipschitz continuous with constant $L_w>0$, and has Lipschitz gradient with constant $L_{\nabla_{\alpha}w}> 0$.    
  \item $\left \| \nabla_{w\alpha}^2\mathcal{L}_1 \right \|$ is bounded that $\left \| \nabla_{w\alpha}^2\mathcal{L}_1 \right \|\leq C_{\mathcal{L}_1^{w\alpha}}$ for some constant $C_{\mathcal{L}_1^{w\alpha}}>0$.      
\end{enumerate}
\end{assumption}

\subsection{Hypergradient based on Neumann Approximation}
In this subsection, we describe how to use the Neumann series to reformulate the hypergradient in DARTS.


\begin{lemma}
\label{lemma_neumannapprox}
Neumann series \cite{lorraine2020optimizing}: With a matrix $A$ that $\left \| I-A \right \|<1$, $A^{-1}=\sum_{k=0}^{\infty}(I-A)^k$.
\end{lemma}

Based on Lemma \ref{lemma_neumannapprox}, the Eq. \eqref{eq:ift_hypergradient} for the IFT based DARTS is formulated by Eq. \eqref{eq:darts_neuman} as described in Corollary \ref{corollary_neumann_hypergradient}.


\begin{corollary}
\label{corollary_neumann_hypergradient}
With small enough learning rate $\gamma<\frac{1}{L^{\nabla_w}_1}$, the hypergradient in DARTS can be formulated as:
\begin{equation} \label{eq:darts_neuman}
 \begin{aligned}
&\nabla_{\alpha}\mathcal{L}_2=\frac{\partial \mathcal{L}_2}{\partial \alpha}-\frac{\partial \mathcal{L}_2}{\partial w}\left [ \frac{\partial^2 \mathcal{L}_1}{\partial w\partial w} \right ]^{-1}\frac{\partial^2 \mathcal{L}_1}{\partial \alpha \partial w}\\
&=\frac{\partial \mathcal{L}_2}{\partial \alpha}-\gamma \frac{\partial \mathcal{L}_2}{\partial w}\sum_{j=0}^{\infty}\left [ I- \gamma \frac{\partial^2 \mathcal{L}_1}{\partial w\partial w} \right ]^j \frac{\partial^2 \mathcal{L}_1}{\partial \alpha \partial w}.
 \end{aligned}
\end{equation}
\end{corollary}

As shown in the Corollary \ref{corollary_neumann_hypergradient}, the approximated hypergradient for DARTS, denoted by $\nabla_{\alpha}\tilde{\mathcal{L}}_2$ could be obtained by only considering the first $K$ terms of Neumann approximation without calculating the inverse of Hessian \cite{shaban2019truncated,lorraine2020optimizing} as,
\begin{equation} \label{eq:neumann_approx_hypergradient}
\resizebox{.85\linewidth}{!}{$
    \displaystyle
\nabla_{\alpha}\tilde{\mathcal{L}}_2=\frac{\partial \mathcal{L}_2}{\partial \alpha}-\gamma \frac{\partial \mathcal{L}_2}{\partial w}\sum_{k=0}^{K}\left [ I- \gamma \frac{\partial^2 \mathcal{L}_1}{\partial w\partial w} \right ]^k \frac{\partial^2 \mathcal{L}_1}{\partial \alpha \partial w}.
$}
\end{equation}
As shown, we could observe the relationship between the proposed $\nabla_{\alpha}\tilde{\mathcal{L}}_2$ and the hypergradient of DARTS in Eq\eqref{eq:darts_hypergradient}, which is the same as $\nabla_{\alpha}\tilde{\mathcal{L}}_2$ when $K=0$. In the following theorem, we give the error bound between our approximated hypergradient $\nabla_{\alpha}\tilde{\mathcal{L}}_2$ and the exact hypergradient $\nabla_{\alpha}\mathcal{L}_2$.

\begin{theorem}
\label{theorem1}
Suppose the inner optimization function $\mathcal{L}_1$ is twice differentiable and is $\mu$-strongly convex with $w$ around $w^*(\alpha)$. The error between the approximated gradient $\nabla_{\alpha}\tilde{\mathcal{L}}_2$ and $\nabla_{\alpha}\mathcal{L}_2$ in DARTS is bounded with $\left \|\nabla_{\alpha}\mathcal{L}_2 -\nabla_{\alpha}\tilde{\mathcal{L}}_2  \right \| \leqslant  C_{\mathcal{L}_1^{w\alpha}}\ C_{\mathcal{L}_2^w}\ \frac{1}{\mu }(1-\gamma \mu )^{K+1}$.
\end{theorem}

Theorem \ref{theorem1} states that the approximated hypergradient approaches to the exact hypergradient as $K$ increases. As described, the form of $\nabla_{\alpha}\tilde{\mathcal{L}}_2$ is  similar to the $K$-step truncated back-propagation in Eq. \eqref{eq:reverse_k_step}, while the memory consumption of our $\nabla_{\alpha}\tilde{\mathcal{L}}_2$ is only $\frac{1}{K}$ of the memory needed to compute $h_{T-K}$, as we only store the gradients of the final solution $w^*$. 
In the following corollary, we describe the connection between the proposed approximated hypergradient $\nabla_{\alpha}\tilde{\mathcal{L}}_2$ and the approximation based on the truncated back-propagation  $h_{T-K}$ \cite{shaban2019truncated}.

\begin{corollary}
\label{corollary2}
When we assume $w_t$ has converged to a stationary point $w^*$ in the last $K$ steps, the proposed $\nabla_{\alpha}\tilde{\mathcal{L}}_2$ is the same as the truncated back-propagation $h_{T-K}$.
\end{corollary}


\subsection{Stochastic Approximation of Hypergradient}

The Lemma \ref{lemma_ksep} and Corollary \ref{corollary2} show that the approximated hypergradient $\nabla_{\alpha}\tilde{\mathcal{L}}_2$ has the potential to be a sufficient descent direction. However, it is not easy to calculate the implicit gradients for DARTS based on Eq. \eqref{eq:neumann_approx_hypergradient} as it needs to deal with large-scale datasets in which the loss functions are large sums of error terms:
\begin{equation} \nonumber
\mathcal{L}_2=\frac{1}{R}\sum_{i=1}^{R}\mathcal{L}_2^i;\qquad \mathcal{L}_1=\frac{1}{J}\sum_{j=1}^{J}\mathcal{L}_1^j,
\end{equation}
where $J$ is the number of minibatches of the training dataset $\mathcal{D}_{train}$ for the inner supernet training $\mathcal{L}_1$, and $R$ is the number of minibatches of the validation dataset $\mathcal{D}_{val}$ for the outer architecture optimization $\mathcal{L}_2$. It is apparently challenging to calculate the gradient based on the full dataset in each step. We therefore utilize  the stochastic gradient based on individual minibatches in practice. That is, we consider the following stochastic approximated hypergradient,
\begin{equation} \label{eq:stoch_neumann_approx_hypergradient}
\resizebox{.85\linewidth}{!}{$
    \displaystyle
\nabla_{\alpha}\hat{\mathcal{L}}_2^i(w^j(\alpha),\alpha)=\frac{\partial \mathcal{L}_2^i}{\partial \alpha}-\gamma \frac{\partial \mathcal{L}_2^i}{\partial w}\sum_{k=0}^{K}\left [ I- \gamma \frac{\partial^2 \mathcal{L}_1^j}{\partial w\partial w} \right ]^k \frac{\partial^2 \mathcal{L}_1^j}{\partial \alpha \partial w}.
$}
\end{equation}
where $\mathcal{L}^i_2$ and $\mathcal{L}^j_1$ correspond to loss functions calculated by randomly sampled minibatches $i$ and $j$ from $\mathcal{D}_{val}$ and $\mathcal{D}_{train}$,  respectively.
\reza{This expression can be computed using the Hessian-vector product technique without explicitly computing the Hessian matrix (see Appendix B). }
Before analyzing the convergence of the proposed $\nabla_{\alpha}\hat{\mathcal{L}}_2(w^j(\alpha),\alpha)$, we give the following lemma to show function $\mathcal{L}_2:\alpha \rightarrow \mathcal{L}_2 (w,\alpha)$ is differentiable with a Lipschitz continuous gradient \cite{couellan2016convergence}.

\begin{lemma}
\label{lemma_Lipschitz_differentiable}
Based on the Assumption \ref{assumption1} and \ref{assumption2}, we have the function $\mathcal{L}_2:\alpha \rightarrow \mathcal{L}_2 (w,\alpha)$ is differentiable with Lipschitz continuous gradient and Lipschitz constant $L_{\nabla_\alpha \mathcal{L}_2}=L^{\nabla_\alpha}_2 + L^{\nabla_w}_2 L_w^2 + L^{w}_2 L_{\nabla_{\alpha}w}$.
\end{lemma}


Then we state and prove the main convergence theorem for the proposed stochastic approximated hypergradient $\nabla_{\alpha}\hat{\mathcal{L}}_2^{i}(w^j(\alpha),\alpha)$ for the differentiable NAS.

\begin{theorem}
\label{theorem2}
Based on several assumptions, we could prove the convergence of the proposed stochastic approximated hypergradient for differentiable NAS. Suppose that: 
\begin{enumerate}
  \item All assumptions in Assumption \ref{assumption1} and \ref{assumption2} and Corollary \ref{corollary_neumann_hypergradient} are satisfied;
  \item $\exists D>0$ such that $E\left [ \left \|  \varepsilon \right \|^2 \right ]\leq D\left \| \nabla_{\alpha}\mathcal{L}_2 \right \|^2$;
  \item $\forall i >0$, $\gamma_{\alpha_i}$ satisfies $\sum_{i=1}^{\infty }\gamma_{\alpha_i}=\infty \qquad$ and  $\qquad \sum_{i=1}^{\infty }\gamma_{\alpha_i}^2< \infty$.
  \item  The inner  function $\mathcal{L}_1$ has the special structure: $\mathcal{L}_1^j(w,\alpha)=h(w,\alpha)+h_j(w,\alpha), \ \ \ \forall j\in 1,...,J,$ that $h_j$ is a linear function with respect to $w$ and $\alpha$.
\end{enumerate} 
  
With small enough learning rate $\gamma_{\alpha}$ for the architecture optimization, the proposed stochastic hypergradient based algorithm converges in expectation to a stationary point, i.e. $\underset{m\rightarrow \infty}{\textup{lim}} E\left [ \left \| \nabla_{\alpha}\hat{\mathcal{L}}_2^{i}(w^j(\alpha_m),\alpha_m) \right \| \right ]=0$.

\end{theorem}

The $\varepsilon$ is defined as the noise term between the stochastic gradient $\nabla_{\alpha}\mathcal{L}_2^{i}(w^j(\alpha),\alpha)$ and the true gradient $\nabla_{\alpha}\mathcal{L}_2$ as:
\begin{equation} \nonumber
\varepsilon_{i,j}=\nabla_{\alpha}\mathcal{L}_2-\nabla_{\alpha}\mathcal{L}_2^{i}(w^j(\alpha),\alpha).
\end{equation}
\reza{where $\nabla_{\alpha}\mathcal{L}_2^{i}(w^j(\alpha),\alpha)$ is the non-approximate version of Eq. \eqref{eq:stoch_neumann_approx_hypergradient} when $K \rightarrow \infty$.}

Theorem \ref{theorem2} shows that the proposed stochastic approximated hypergradient is also a sufficient descent direction, which leads the differentiable NAS converges to a stationary point. The conditions 2-4 in Theorem \ref{theorem2} are common assumptions in analyzing the stochastic bi-level gradient methods \cite{couellan2016convergence,ghadimi2018approximation,grazzi2020convergence,grazzi2020iteration}. We assume that $\mathcal{L}_1$ in Eq. \eqref{eq:darts_neuman} is $\mu$-strongly convex with $w$ around $w^*$, which can be made possible by appropriate choice of learning rates \cite{rajeswaran2019meta,shaban2019truncated}. Another key assumption in our convergence analysis is the Lipshitz differentiable assumptions for $\mathcal{L}_1$ and $\mathcal{L}_2$ in Assumption \ref{assumption1} and \ref{assumption2}, which also received considerable attention in recent optimization and deep learning literature \cite{jin2017escape,rajeswaran2019meta,lorraine2020optimizing,mackay2018self,grazzi2020iteration}. 


\begin{algorithm}[t]
\caption{iDARTS}
\label{alg:algorithm1}
\textbf{Input}: $\mathcal{D}_{train}$ and $\mathcal{D}_{val}$. Initialized supernet weights $w$ and operations magnitude $\alpha_\theta$. 
\begin{algorithmic}[2]
\WHILE{\textit{not converged}}
\STATE $\star$ Sample batches from $\mathcal{D}_{train}$. Update supernet weights $w$ based on cross-entropy loss with $T$ steps.
\STATE $\star$ Get the Hessian matrix $\frac{\partial^2 \mathcal{L}_1}{\partial w\partial w}$. 
\STATE $\star$ Sample batch from $\mathcal{D}_{val}$. Calculate hypergradient $\nabla_{\alpha}\hat{\mathcal{L}}_2^i(w^j(\alpha),\alpha)$ based on Eq.\eqref{eq:stoch_neumann_approx_hypergradient}, and update $\alpha$ with $\alpha \leftarrow \alpha-\gamma_\alpha \nabla_{\alpha}\hat{\mathcal{L}}_2^i(w^j(\alpha),\alpha)$.
\ENDWHILE
\STATE Obtain $\alpha^*$ through \textbf{argmax}.
\end{algorithmic}
\end{algorithm}

\subsection{Differentiable Architecture Search with Stochastic Implicit Gradients}

Different from DARTS that alternatively optimizes both $\alpha$ and $w$ with only one step in each round, iDARTS is supposed to train the supernet with enough steps to make sure the $w(\alpha)$ is near $w^*(\alpha)$ before optimizing $\alpha$. The framework of our iDARTS is sketched in Algorithm 1. Generally, it is impossible to consider a very large $T$ for each round of supernet weights $w$ optimization, as the computational cost increase linear with $T$. Fortunately, empirical experiments show that, with the weight sharing, the differentiable NAS can adapt $w(\alpha)$ to $w^*(\alpha)$ with a small $T$ in the later phase of architecture search. 



\section{Experiments}


\begin{table*}
\footnotesize
\centering
\caption{Comparison results with NAS baselines on NAS-Bench-201. 
}
\begin{tabular}
{|l|c|c|c|c|c|c|c|c|}
\hline

\makecell[c]{\multirow{2}*{Method}}&\multicolumn{2}{c|}{CIFAR-10}&\multicolumn{2}{c|}{CIFAR-100}&\multicolumn{2}{c|}{ImageNet-16-120}\\
~&\multicolumn{1}{c}{Valid(\%)}&\multicolumn{1}{c|}{Test(\%)}&\multicolumn{1}{c}{Valid(\%)}&\multicolumn{1}{c|}{Test(\%) }&\multicolumn{1}{c}{Valid(\%) }&\multicolumn{1}{c|}{Test(\%)}\\
\hline
\hline
ENAS \cite{pham2018efficient}&37.51$\pm$3.19&53.89$\pm$0.58&13.37$\pm$2.35&13.96$\pm$2.33&15.06$\pm$1.95&14.84$\pm$2.10\\
RandomNAS \cite{li2019random}&80.42$\pm$3.58&84.07$\pm$3.61&52.12$\pm$5.55&52.31$\pm$5.77&27.22$\pm$3.24&26.28$\pm$3.09\\
SETN \cite{dong2019one}&84.04$\pm$0.28&87.64$\pm$0.00&58.86$\pm$0.06&59.05$\pm$0.24&33.06$\pm$0.02&32.52$\pm$0.21\\
GDAS \cite{GDAS}&89.88$\pm$0.33&93.40$\pm$0.49&70.95$\pm$0.78&70.33$\pm$0.87&41.28$\pm$0.46&41.47$\pm$0.21\\
DARTS \cite{liu2018darts}&39.77$\pm$0.00&54.30$\pm$0.00&15.03$\pm$0.00&15.61$\pm$0.00&16.43$\pm$0.00&16.32$\pm$0.00\\
\hline
iDARTS &89.86$\pm$0.60&93.58$\pm$0.32&70.57$\pm$0.24&70.83$\pm$0.48&40.38$\pm$0.593&40.89$\pm$0.68\\
\hline
\textbf{optimal}&91.61&94.37&74.49&73.51&46.77&47.31\\
\hline
\end{tabular}
\flushleft{iDARTS's best single run achieves \textbf{93.76\%}, \textbf{71.11\%}, and \textbf{41.44\%} test accuracy on CIFAR-10, CIFAR-100, and ImageNet, respectively. }
\label{tab:nasbench201}
\end{table*}

In Section \ref{sec3}, we have theoretically shown that our iDARTS can asymptotically compute the exact hypergradient and lead to a convergence in expectation to a stationary point for the architecture optimization. In this section, we conduct a series of experiments to verify whether the iDARTS leads to better results in the differentiable NAS with realistic settings. We consider three different cases to analyze iDARTS, including two NAS benchmark datasets, NAS-Bench-1Shot1 \cite{zela2020nasbench1shot1} and NAS-Bench-201 \cite{BENCH102}, and the common DARTS search space \cite{liu2018darts}. We first analyze the iDARTS on the two NAS benchmark datasets, along with discussions of hyperparameter settings. Then we compare iDARTS with state-of-the-art NAS methods on the common DARTS search space.

\begin{figure}
 \subfloat[Validation error]{
  \begin{minipage}{4.2cm}
      \includegraphics[width=4.2cm,height=3cm]{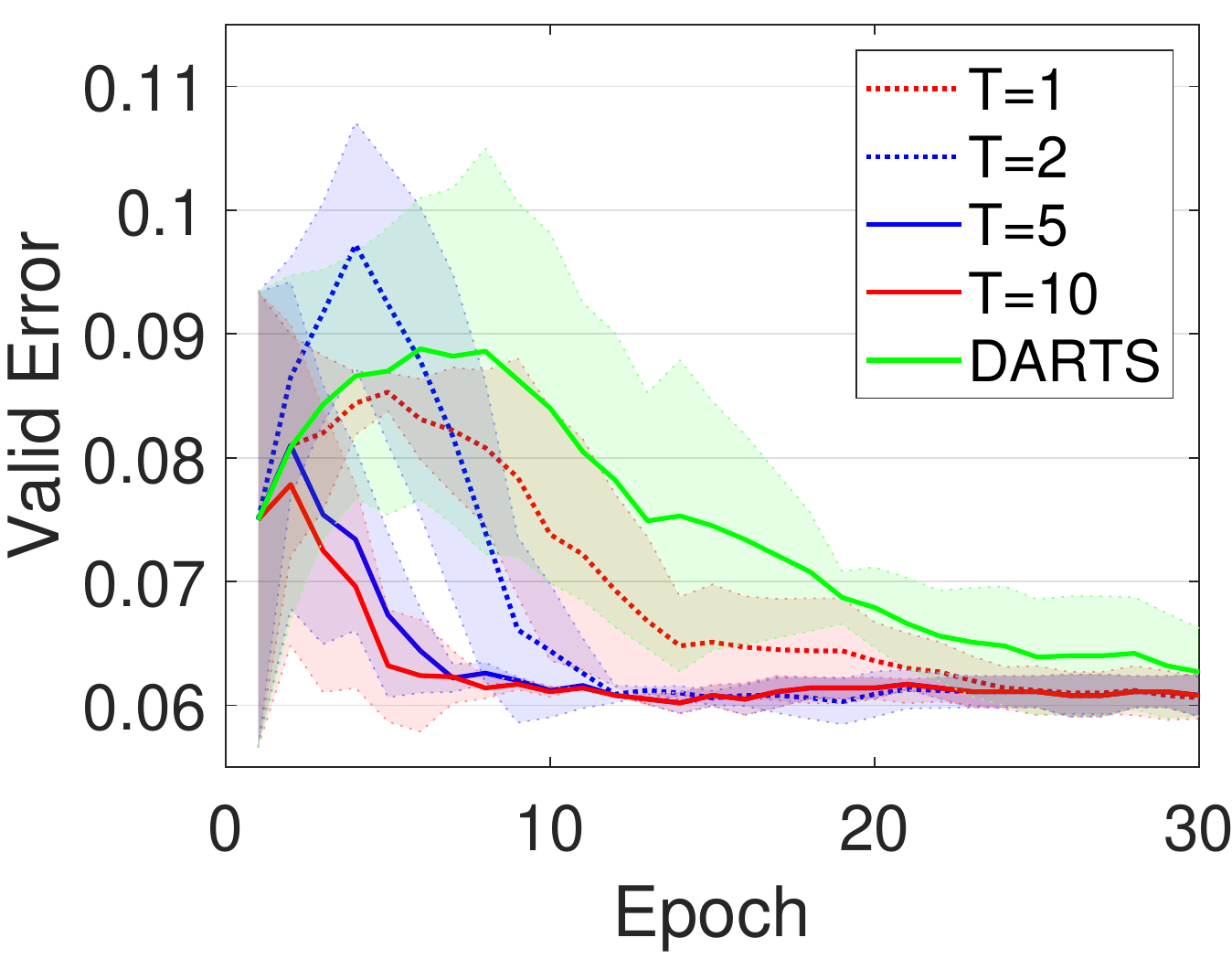}
  \end{minipage}
 }
  \subfloat[Test errors]{
  \begin{minipage}{4.2cm}
       \includegraphics[width=4.2cm,height=3cm]{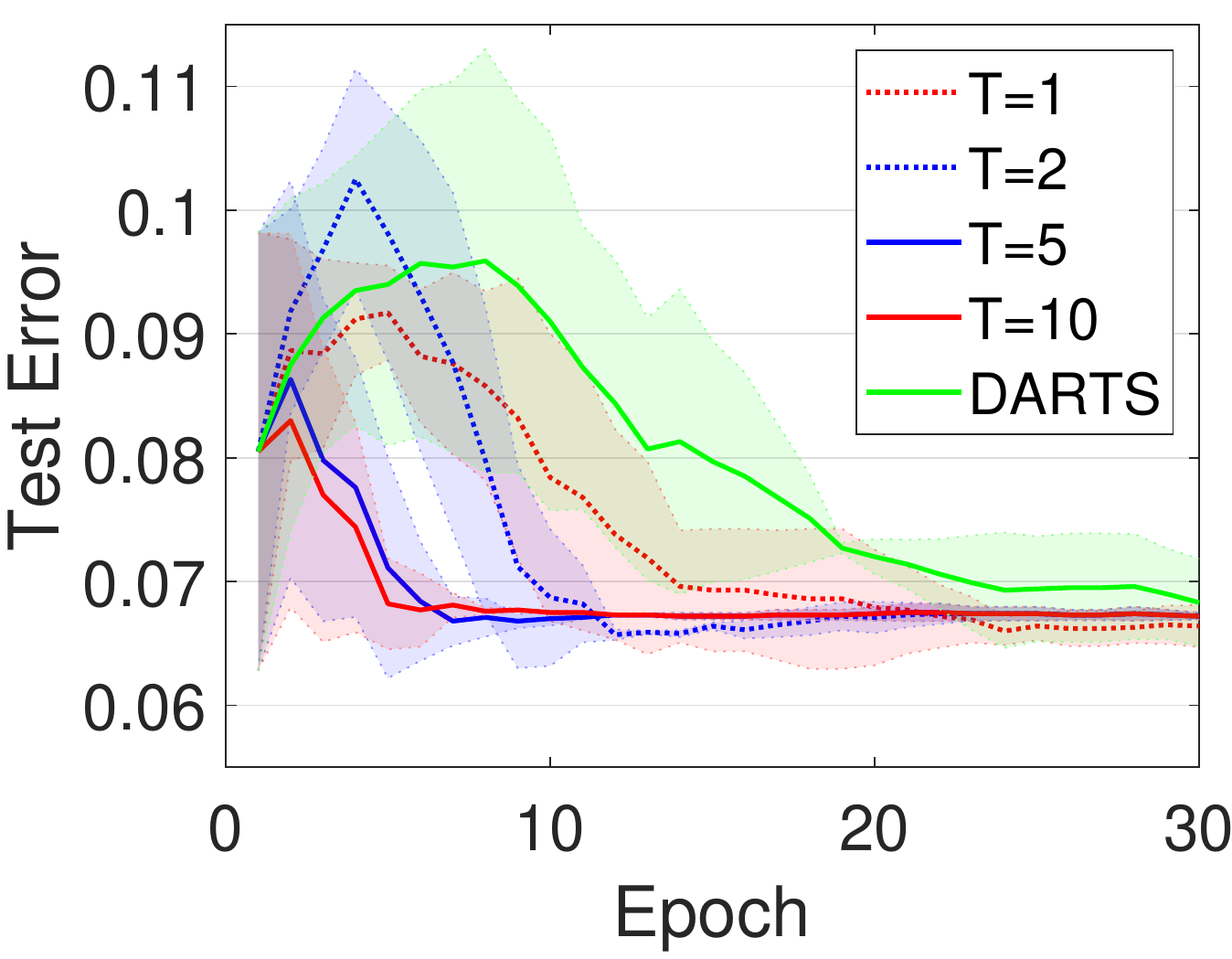}
  \end{minipage} 
  }
  
 \caption{Validation and test errors of iDARTS with different $T$ and DARTS on the search space 3 of NAS-Bench-1Shot1.}
 \label{fig:T_nas1shot1}
\end{figure}

\subsection{Reproducible Comparison on NAS-Bench-1Shot1}
\label{sec5.1}


To study the empirical performance of iDARTS, we run the iDARTS on the NAS-Bench-1Shot1 dataset with different \textit{random seeds} to report its statistical results, and compare with the most closely related baseline DARTS \cite{liu2018darts}. The NAS-Bench-1Shot1 is built from the NAS-Bench-101 benchmark dataset \cite{ying2019bench}, through dividing all architectures in NAS-Bench-101 into 3 different unified cell-based search spaces. The architectures in each search space have the same number of nodes and connections, making the differentiable NAS could be directly applied to each search space. The three search spaces contain 6240, 29160, and 363648 architectures with the CIFAR-10 performance, respectively. We choose the third search space in NAS-Bench-1Shot1 to analyse iDARTS, since it is much more complicated than the remaining two search spaces and is a better case to identify the advantages of iDARTS. 

Figure \ref{fig:T_nas1shot1} plots the mean and standard deviation of the validation and test errors for iDARTS and DARTS, with tracking the performance during the architecture search on the NAS-Bench-1Shot1 dataset. As shown, our iDARTS with different $T$ generally outperforms DARTS during the architecture search in terms of both validation and test error. More specifically, our iDARTS significantly outperforms the baseline in the early stage, demonstrating that our iDARTS could quickly find superior architectures and is more stable.

As described, one significant difference from DARTS is that iDARTS can conduct more than one training step in the inner-loop optimization. Figure \ref{fig:T_nas1shot1} also analyzes the effects of the inner optimization steps $T$, plotting the performance of iDARTS with different $T$ on the NAS-Bench-1Shot1. As shown, the inner optimization steps positively affect the performance of iDARTS, where increasing $T$ helps iDARTS converge to excellent solutions faster. One underlying reason is that increasing $T$ could adapt $w$ to a local optimal $w^*$, thus helping iDRTS approximate the exact hypergradient more accurately. We should notice that the computational cost of iDARTS also increases with $T$, and our empirical findings suggest a $T=5$ achieves an excellent compute and performance trade-off for iDARTS on NAS-Bench-1shot1. More interesting, iDARTS with $T=1$ is similar as DARTS which both conduct the inner optimization with only one step, with the difference that iDARTS adopts the Neumann approximation while DARTS considers the unrolled differentiation. We could observe that iDARTS still outperforms DARTS by large margins in this case, showing the superiority of the proposed approximation over DARTS.


\begin{figure}[ht]
 \subfloat[Validation and test performance with different $T$]{
  \begin{minipage}{4.2cm}
      \includegraphics[width=4.1cm,height=3cm]{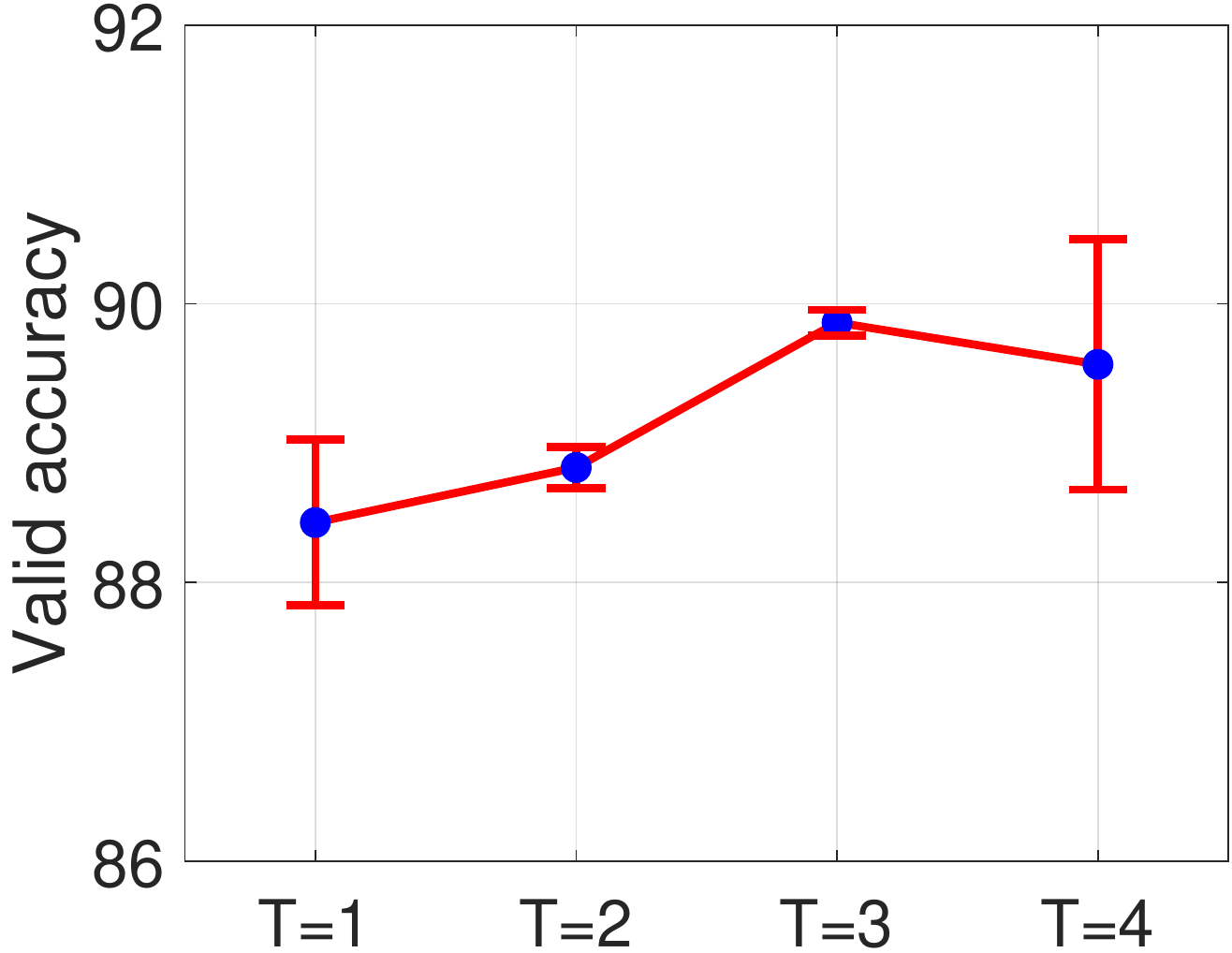}
  \end{minipage}
  \begin{minipage}{4.2cm}
       \includegraphics[width=4.1cm,height=3cm]{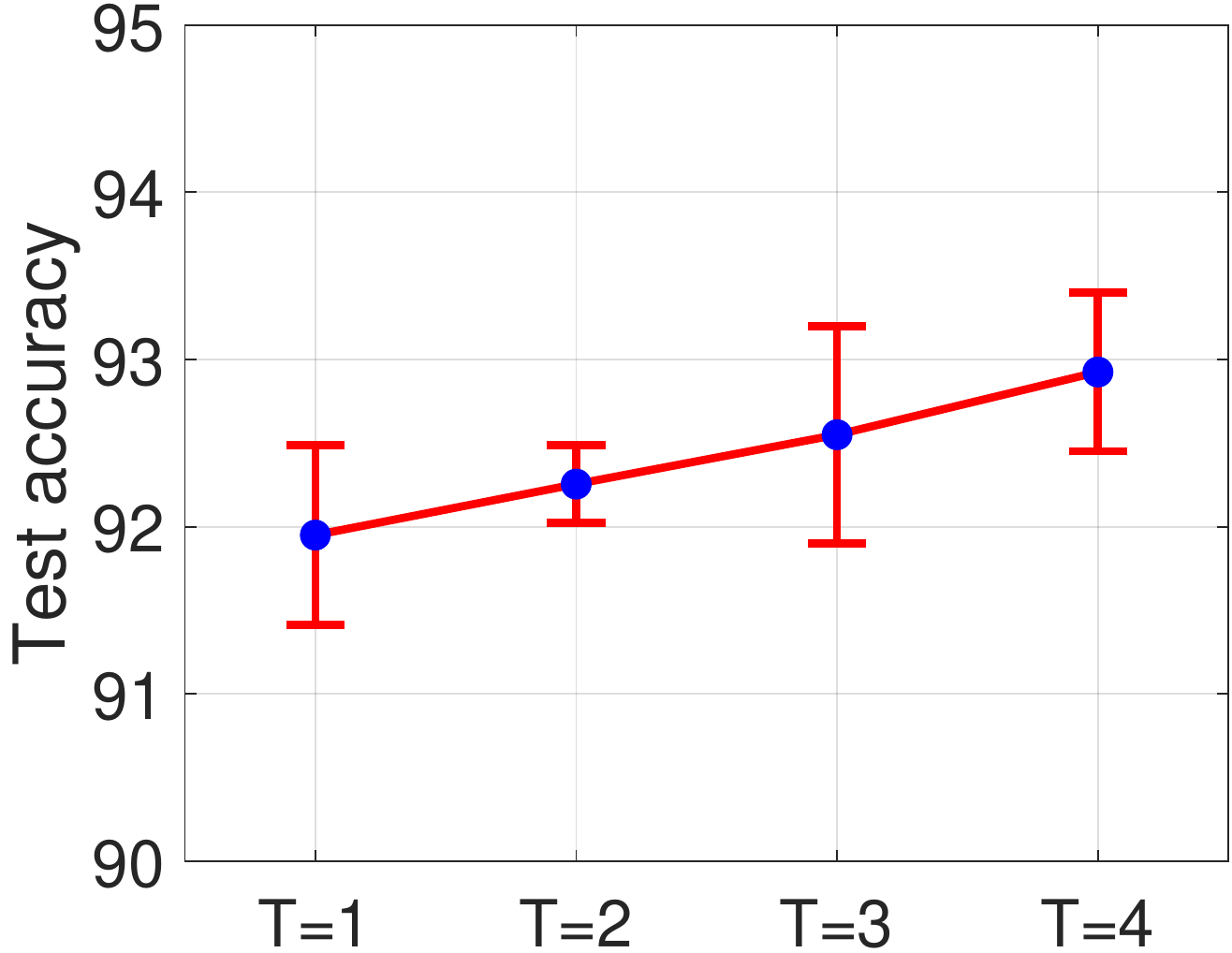}
  \end{minipage} 
  }
  
 \subfloat[Validation and test performance with different $\gamma$]{
  \begin{minipage}{4.2cm}
      \includegraphics[width=4.1cm,height=3cm]{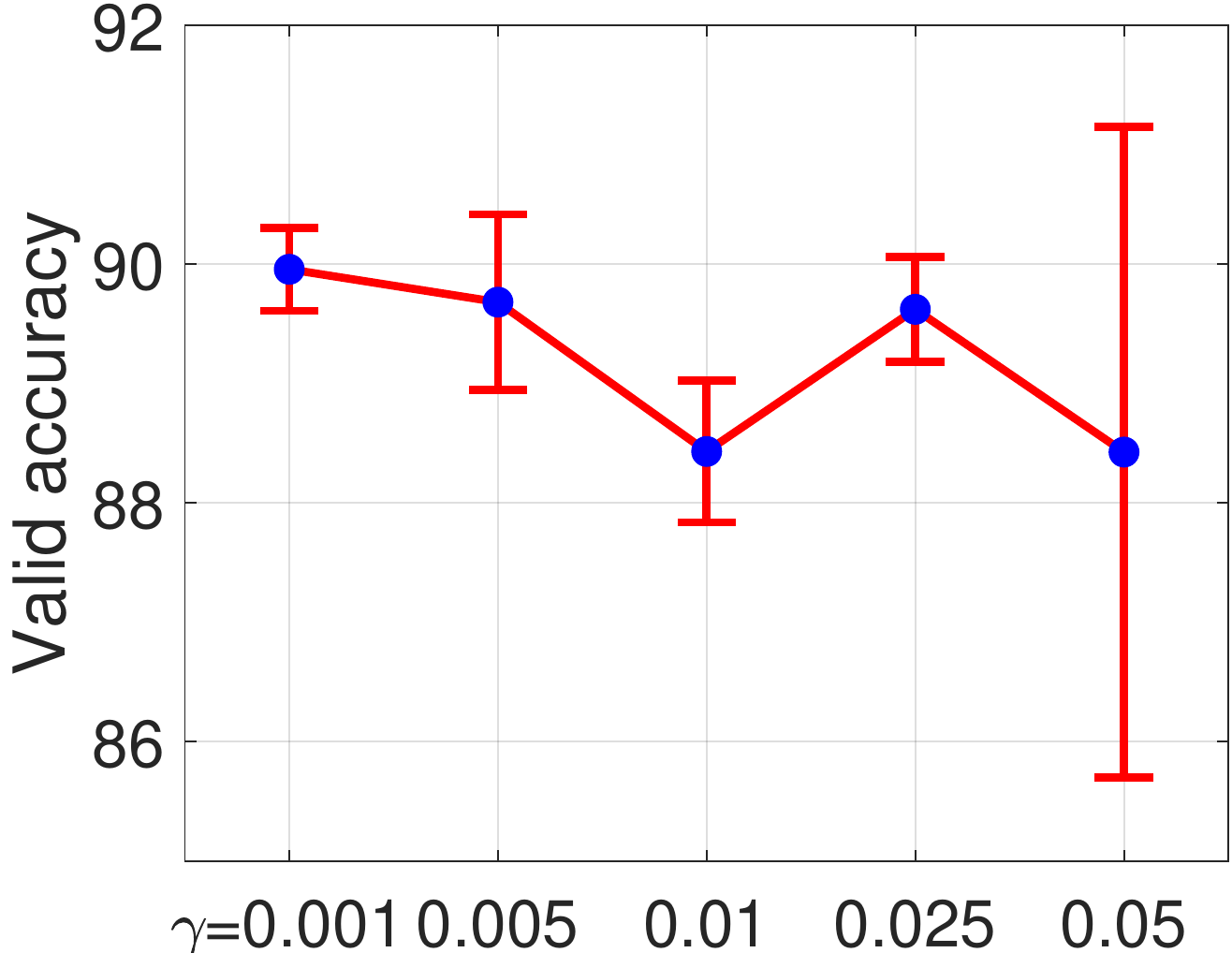}
  \end{minipage}
  \begin{minipage}{4.2cm}
       \includegraphics[width=4.1cm,height=3cm]{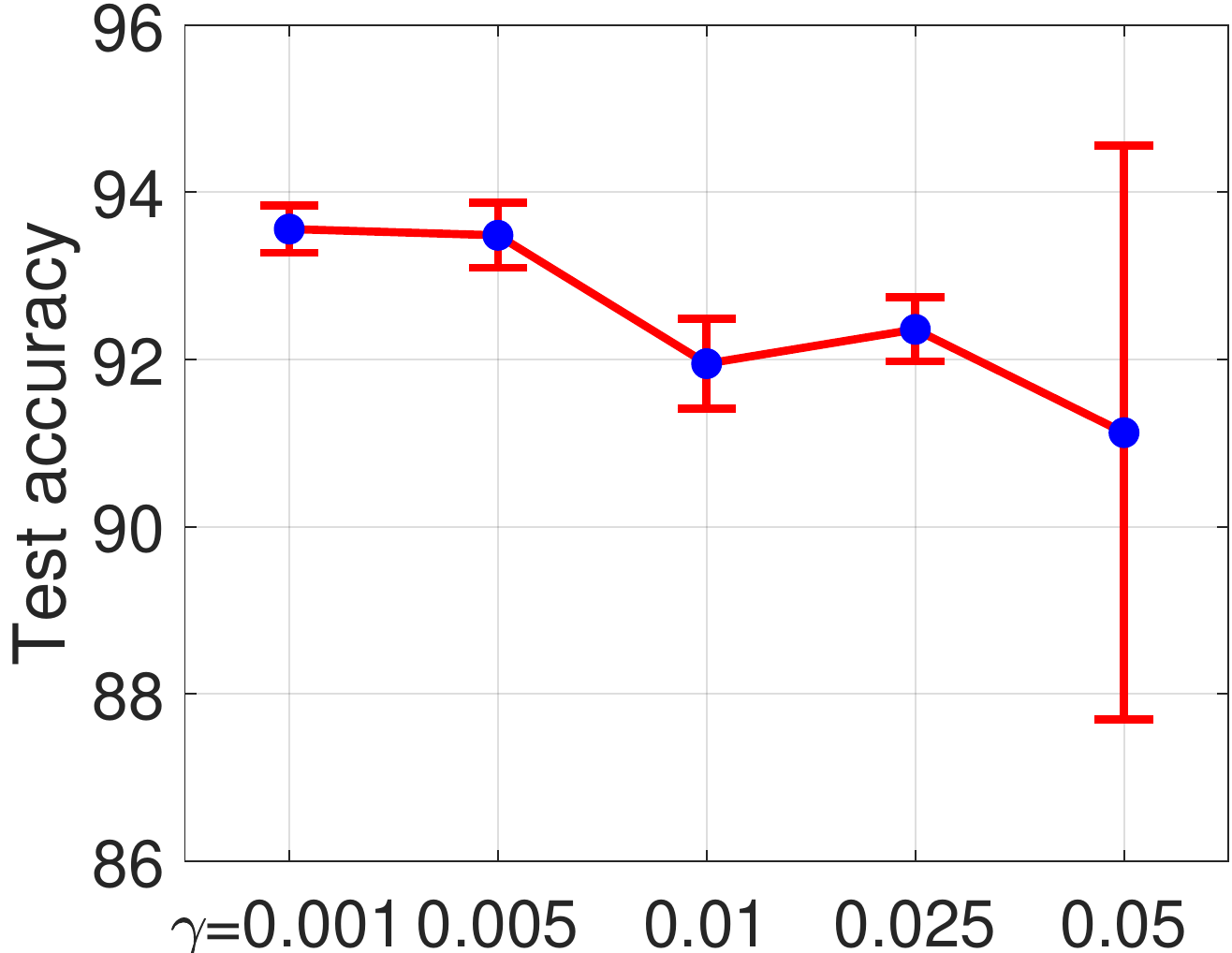}
  \end{minipage} 
  }
  
 \subfloat[Validation and test performance with different $\gamma_\alpha$]{
  \begin{minipage}{4.2cm}
      \includegraphics[width=4.1cm,height=3cm]{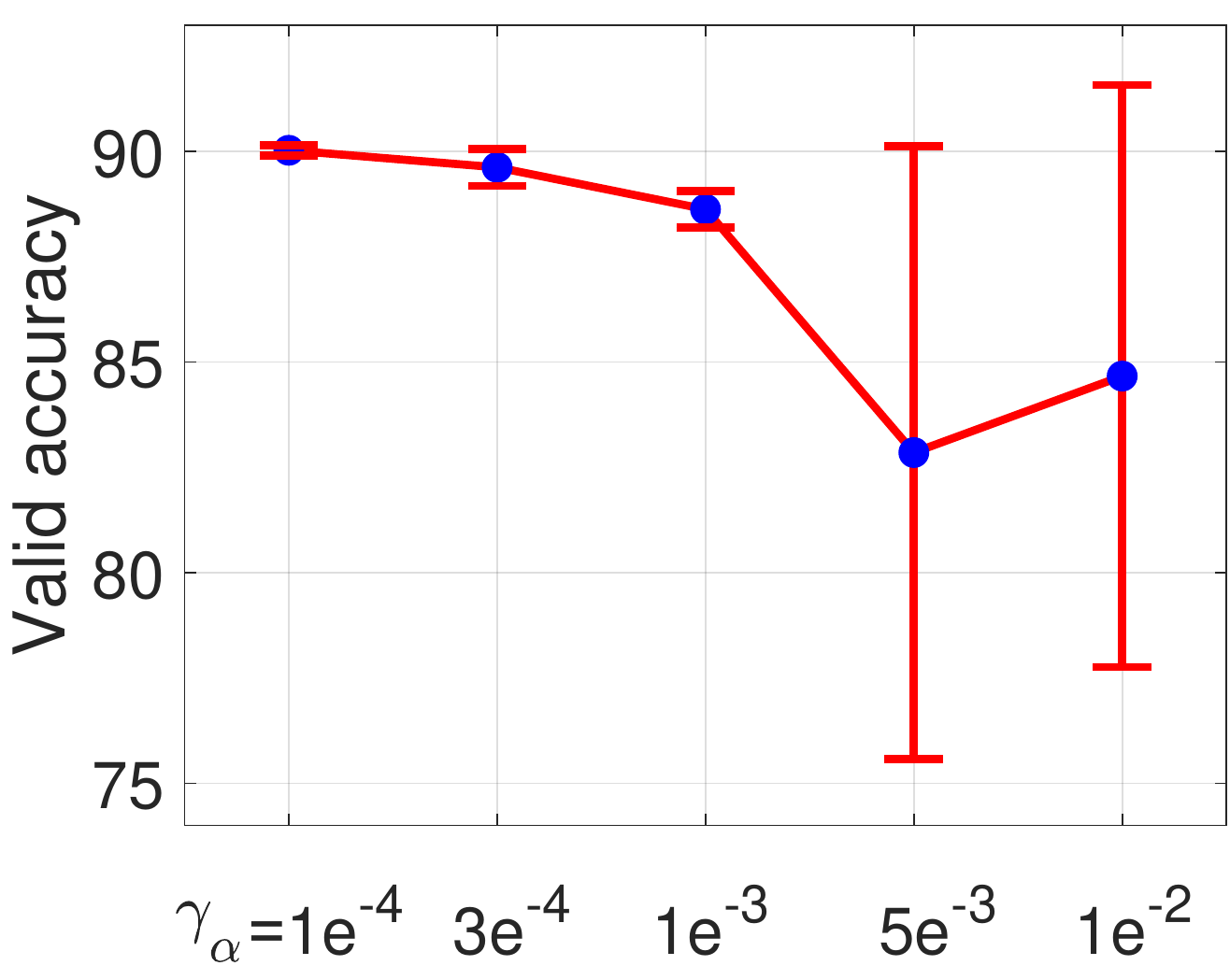}
  \end{minipage}
  \begin{minipage}{4.2cm}
       \includegraphics[width=4.1cm,height=3cm]{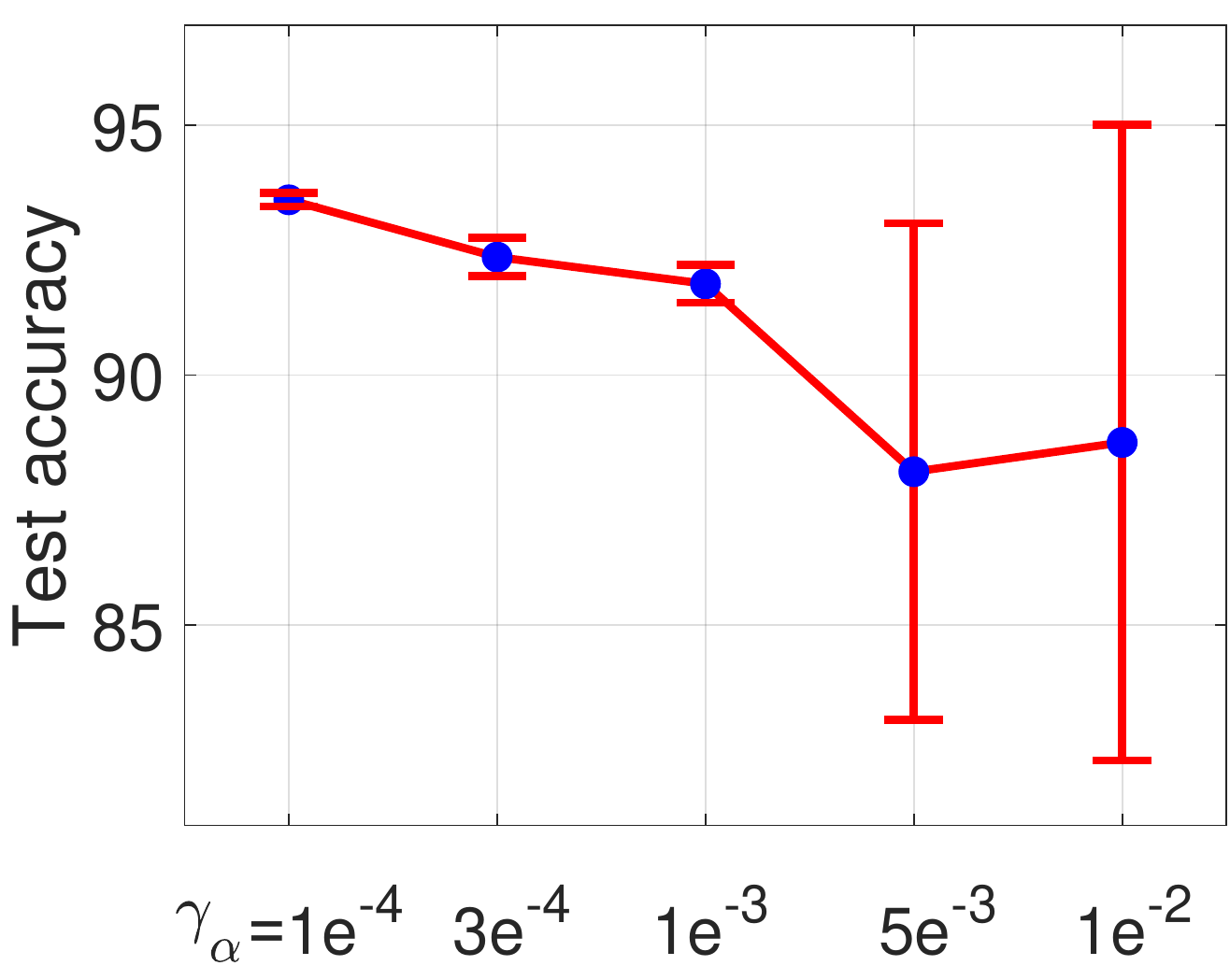}
  \end{minipage} 
  }  
  
 \caption{Hyperparameter analysis of iDARTS on the NAS-Bench-201 benchmark dataset. }
 \label{fig:T_nas201}
\end{figure}

\subsection{Reproducible Comparison on NAS-Bench-201}
\label{sec5.2}

The NAS-Bench-201 dataset \cite{BENCH102} is another popular NAS benchmark dataset to analyze differentiable NAS methods. The search space in NAS-Bench-201 contains four nodes with five associated operations, resulting in 15,625 cell candidates. The search space of NAS-Bench-201 is much simpler than NAS-Bench-1Shot1, while it contains the performance of CIFAR-100, CIFAR-100, and ImageNet for all architectures in this search space.

Table \ref{tab:nasbench201} summarizes the performance of iDARTS on NAS-Bench-201 compared with differentiable NAS baselines, where the statistical results are obtained from independent search experiments with different \textit{random seeds}. As shown, our iDARTS achieved excellent results on the NAS-Bench-201 benchmark and significantly outperformed the DARTS baseline, with a 93.76\%, 71.11\%, and 41.44\% test accuracy on CIFAR-10, CIFAR-100, and ImageNet, respectively. As described in Section \ref{sec3}, iDARTS is built based on the DARTS framework, with only reformulating the hypergradient calculation. These results in Table \ref{tab:nasbench201} verified the effectiveness of our iDARTS, which outperforms DARTS by large margins. 


Similar to the experiments in the NAS-Bench-1Shot1, we also analyze the importance of hyperparameter $T$ in the NAS-Bench-201 dataset. Figure \ref{fig:T_nas201} (a) summaries the performance of iDARTS with different number of inner optimization steps $T$ on the NAS-Bench-201. As demonstrated, the performance of iDARTS is sensitive to the hyperparameter $T$, and a larger $T$ helps iDARTS to achieve better results while also increases the computational time, which is also in line with the finding in the NAS-Bench-1Shot1. We empirically find that $T=4$ is enough to achieve competitive results on NAS-Bench-201.

The hypergradient calculation of iDARTS is based on the Neumann approximation in Eq.\eqref{eq:darts_neuman}, and one underlying condition is that the learning rates $\gamma$ for the inner optimization should be small enough to make $\left \| I- \gamma \frac{\partial^2 \mathcal{L}_1}{\partial w\partial w} \right \|<1$. We also conduct an ablation study to analyze how this hyperparameter affects our iDARTS, where Figure \ref{fig:T_nas201} (b) plots the performance of iDARTS with different learning rates $\gamma$ for the inner optimization on the NAS-Bench-201. As shown, the performance of iDARTS is sensitive to $\gamma$, and a smaller $\gamma$ is preferred, which also offers support for the Corollary \ref{corollary_neumann_hypergradient} and Theorem \ref{theorem1}.

During the analysis of the convergence of iDARTS, the learning rate $\gamma_{\alpha}$ plays a key role in the hypergradient approximation for the architecture optimization. Figure \ref{fig:T_nas201} (c) also summaries the performance of iDARTS with different initial learning rate $\gamma_{\alpha}$ on the NAS-Bench-201.  As shown in Figure \ref{fig:T_nas201} (c), the performance of iDARTS is sensitive to $\gamma_{\alpha}$, where a smaller $\gamma_{\alpha}$ is recommended, and a large $\gamma_{\alpha}$ is hardly able to converge to a stationary point. An underlying reason may lay in the proof of Theorem \ref{theorem2}, that choosing a small enough $\gamma_{\alpha}$ guarantees that the iDARTS converges to a stationary point.



\subsection{Experiments on DARTS Search Space}
\label{sec5.3}

We also apply iDARTS to a convolutional architecture search in the common DARTS search space \cite{liu2018darts} to compare with the state-of-the-art NAS methods, where all experiment settings are following DARTS for fair comparisons. The search procedure needs to look for two different types of cells on CIFAR-10: normal cell $\alpha_{normal}$ and reduction cell $\alpha_{reduce}$,  to stack more cells to form the final structure for the architecture evaluation. The best-found cell on CIFAR-10 is then transferred to CIFAR-100 and ImageNet datasets to evaluate its transferability. 

\begin{table*}
\centering
\caption{Comparison results with state-of-the-art weight-sharing NAS approaches.}

\begin{tabular}{|l|c|c|c|c|c|c|c|}
\hline

\makecell[c]{\multirow{2}*{Method}}&\multicolumn{3}{c|}{Test Error (\%)}&\multicolumn{1}{c|}{Param}&\multicolumn{1}{c|}{$+ \times$}&\multicolumn{1}{c|}{Architecture}\\
~&\multicolumn{1}{c}{CIFAR-10}&\multicolumn{1}{c}{CIFAR-100}&\multicolumn{1}{c|}{ImageNet}&{(M)}&{(M)}&{Optimization}\\
\hline\hline
NASNet-A \cite{zoph2016neural}&2.65&17.81&26.0 / 8.4&3.3&564&RL\\
PNAS \cite{liu2018progressive}&3.41$\pm$0.09 &17.63&25.8 / 8.1&3.2&588&SMBO\\
AmoebaNet-A \cite{real2018regularized}&3.34$ \pm$0.06&-&25.5 / 8.0&3.2&555&EA\\
ENAS \cite{pham2018efficient}&2.89&18.91&-&4.6&-&RL\\
EN$^2$AS \cite{miaoijcai2020}&2.61$\pm$0.06&16.45&26.7 / 8.9&3.1&506&EA\\
RandomNAS \cite{li2019random}&2.85$\pm$0.08&17.63 &27.1&4.3&613&random\\
NSAS \cite{miaozhang20}&2.59$\pm$0.06&17.56 &25.5 / 8.2 &3.1&506&random\\
\hline
PARSEC \cite{casale2019probabilistic}&2.86$\pm$0.06 &-&26.3&3.6&509&gradient\\
SNAS \cite{xie2018snas}&2.85$\pm$0.02&20.09 &27.3 / 9.2&2.8&474&gradient\\
SETN \cite{dong2019one}&2.69&17.25&25.7 / 8.0&4.6&610&gradient\\
MdeNAS \cite{zheng2019multinomial}&2.55&17.61&25.5 / 7.9&3.6&506&gradient\\
GDAS \cite{GDAS}&2.93&18.38&26.0 /  8.5&3.4&545&gradient\\
XNAS* \cite{nayman2019xnas}&2.57$\pm$0.09&16.34&24.7 / 7.5&3.7&600&gradient\\
PDARTS \cite{chen2019progressive}&2.50&16.63& 24.4 / 7.4&3.4&557&gradient\\
PC-DARTS \cite{xu2019pcdarts}&2.57$\pm$0.07&17.11&25.1 / 7.8&3.6&586&gradient\\
DrNAS \cite{chen2020drnas}&2.54$\pm$0.03&16.30&24.2 / 7.3&4.0&644&gradient\\
DARTS \cite{liu2018darts}&2.76$\pm$0.09&17.54&26.9 / 8.7&3.4&574&gradient\\
\hline\hline
iDARTS&\textbf{2.37$\pm$0.03} & \textbf{16.02}&\textbf{24.3 / 7.3}&3.8&595&gradient\\
\hline
\end{tabular}
\flushleft{``*" indicates the results reproduced based on the best-reported cell structures with a common experimental setting \cite{liu2018darts}. ``Param" is the model size when applied on CIFAR-10, while ``$+\times$" is calculated based on the ImageNet dataset.}

\label{tab:results_CIFAR}
\end{table*}

\begin{figure}
 \subfloat[normal cell]{
  \begin{minipage}{4.2cm}
      \includegraphics[width=4.2cm,height=3cm]{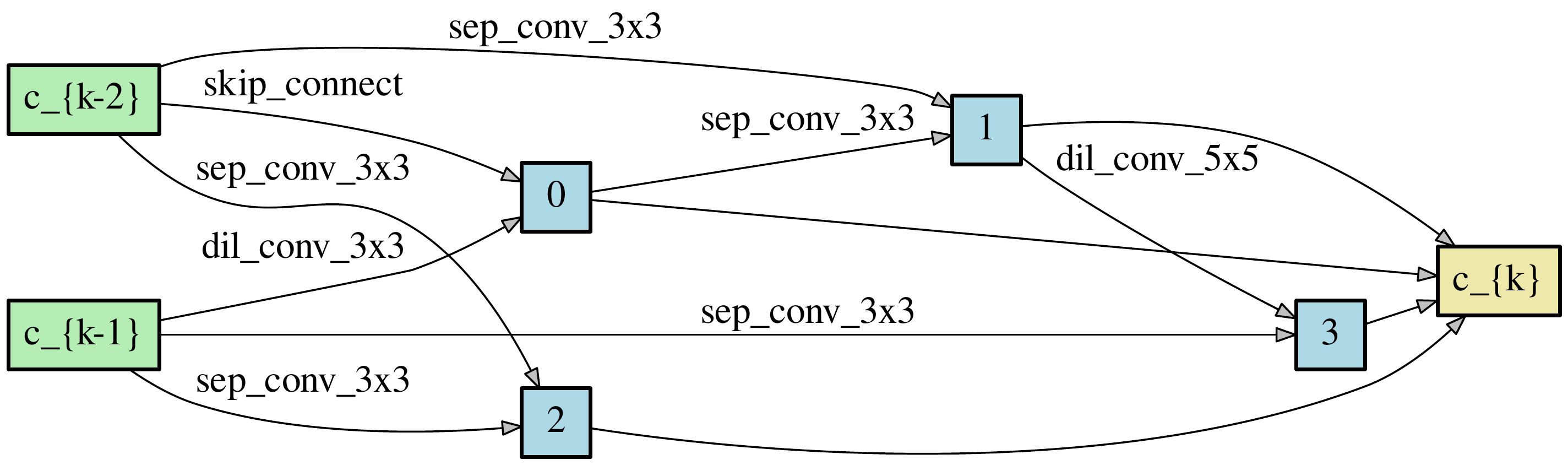}
  \end{minipage}
 }
  \subfloat[reduction cell]{
  \begin{minipage}{4.2cm}
       \includegraphics[width=4.2cm,height=3cm]{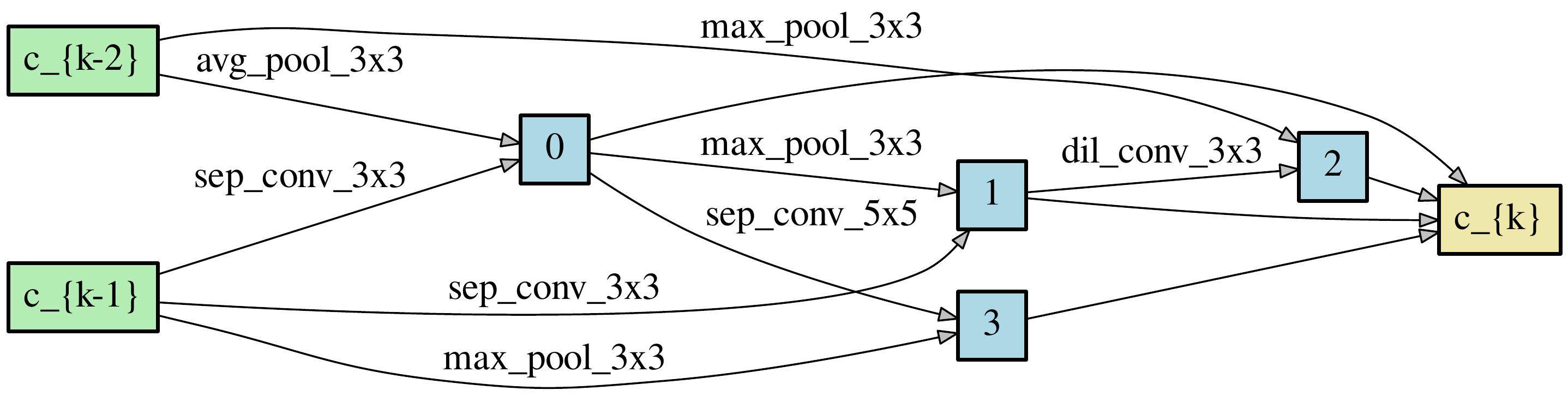}
  \end{minipage} 
  }
  
 \caption{The best cells discovered by iDARTS on the DARTS search space. }
 \label{fig:cellstructure}
\end{figure}

The comparison results with the state-of-the-art NAS methods are presented in Table \ref{tab:results_CIFAR}, and Figure \ref{fig:cellstructure} demonstrates the best-found architectures by iDARTS. As shown in Table \ref{tab:results_CIFAR}, iDARTS achieves a 2.37$\pm$0.03 \% test error on CIFAR-10 (where the best single run is 2.35\%), which is on par with the state-of-the-art NAS methods and outperforms the DARTS baseline by a large margin, again verifying the effectiveness of the proposed method. 

Following DARTS experimental setting, the best-searched architectures on CIFAR-10 are then transferred to CIFAR-100 and ImageNet to evaluate the transferability. The evaluation setting for CIFAR-100 is the same as CIFAR-10. In the ImageNet dataset, the experiment setting is slightly different from CIFAR-10 in that only 14 cells are stacked, and the number of initial channels is changed to 48. We also follow the mobile setting in \cite{liu2018darts} to restrict the number of multiply-add operations (``$+ \times$") to be less than 600M on the ImageNet. The comparison results with state-of-the-art differentiable NAS approaches on CIFAR-100 and ImageNet are demonstrated in Table \ref{tab:results_CIFAR}. As shown, iDARTS delivers a competitive result with 16.02\% test error on the CIFAR-100 dataset, which is a state-of-the-art performance and outperforms peer algorithms by a large margin. On the ImageNet dataset, the best-discovered architecture by our iDARTS also achieves a competitive result with 24.4 / 7.3 \% top1 / top5 test error, outperforming or on par with all peer algorithms. Please note that, although DrNAS achieved outstanding performance on ImageNet, the number of multiply-add operations of its searched model is much over 600 M, violating the mobile-setting.

\section{Conclusion}
This paper opens up a promising research direction for NAS by focusing on the hypergradient approximation in the differentiable NAS. We introduced the implicit function theorem (IFT) to reformulate the hypergradient calculation in the differentiable NAS, making it practical with numerous inner optimization steps. To avoid calculating the inverse of the Hessian matrix, we utilized the Neumann series to approximate the inverse, and further devised a stochastic approximated hypergradient to relieve the computational cost. We theoretically analyzed the convergence and proved that the proposed method, called iDARTS, is expected to converge to a stationary point when applied to a differentiable NAS. We based our framework on DARTS and performed extensive experimental results that verified the proposed framework's effectiveness. While we only considered the proposed stochastic approximated hypergradient for differentiable NAS, iDARTS can in principle be used with a variety of bi-level optimization applications, including in meta-learning and hyperparameter optimization, opening up several interesting avenues for future research.



\section*{Acknowledgement}
This study was supported in part by the Australian Research Council (ARC) under a Discovery Early Career Researcher Award (DECRA) No. DE190100626 and DARPA’s Learning with Less Labeling (LwLL) program under agreement FA8750-19-2-0501.

{\small
\bibliographystyle{icml2020}
\bibliography{example_paper}
}

\onecolumn{
\section*{A. Proof}

\newenvironment{proof_l1}{\paragraph{Proof of Lemma \ref{lemma_ift}:}}{  $\square$}
\begin{proof_l1}
Based on the implicit function theorem \cite{lorraine2020optimizing}, or we simply set $\frac{\mathcal{L}_1(w^*,\alpha)}{\partial w}=0$ since the model weights $w$ achieved the local optimal in the training set with $\alpha$, we have:
\begin{equation} \label{eq:loc_opt}
\frac{\partial \mathcal{L}_1(w^*(\alpha), \alpha)}{\partial w}=0,
\end{equation}
and we have
\begin{equation} 
\begin{aligned}
\frac{\partial }{\partial \alpha}\left ( \frac{\partial \mathcal{L}_1(w^*(\alpha), \alpha)}{\partial w} \right )=0, \\
\frac{\partial^2 \mathcal{L}_1}{\partial \alpha \partial w}+ \frac{\partial^2 \mathcal{L}_1}{\partial w\partial w}\frac{\partial (w^*(\alpha))}{\partial \alpha}=0,\\
\frac{\partial (w^*(\alpha))}{\partial \alpha}=-\left [ \frac{\partial^2 \mathcal{L}_1}{\partial w\partial w} \right ]^{-1}\frac{\partial^2 \mathcal{L}_1}{\partial \alpha \partial w}.
\end{aligned}
\end{equation}
In this way, the hypergradient could be formulated as
\begin{equation} 
\nabla_{\alpha}\mathcal{L}_2=\frac{\partial \mathcal{L}_2}{\partial \alpha}-\frac{\partial \mathcal{L}_2}{\partial w}\left [ \frac{\partial^2 \mathcal{L}_1}{\partial w\partial w} \right ]^{-1}\frac{\partial^2 \mathcal{L}_1}{\partial \alpha \partial w}.
\end{equation}
\end{proof_l1}

\newenvironment{proof3}{\paragraph{Proof of Corollary \ref{corollary_neumann_hypergradient}:}}{  $\square$}
\begin{proof3}
The key in this proposition is to use the Neumann series to approximate the $\left [ \frac{\partial^2 \mathcal{L}_1}{\partial w\partial w} \right ]^{-1}$. 


Based on the Neumann series approximation, for $\left \| I-A \right \|<1$, we have:
\begin{equation} \label{eq:Neumann_series_2}
A^{-1}=\sum_{k=0}^{\infty}(I-A)^k.
\end{equation}



Based Assumption \ref{assumption2}.1, we have $\frac{\partial^2 \mathcal{L}_1}{\partial w\partial w}<L^{\nabla_w}_1$. With $\gamma<\frac{1}{L^{\nabla_w}_1}$, we have $\left \| I- \gamma \frac{\partial^2 \mathcal{L}_1}{\partial w\partial w} \right \|<1$ \cite{shaban2019truncated,lorraine2020optimizing}. When we conduct the Neumann series approximation for $\left [ \frac{\partial^2 \mathcal{L}_1}{\partial w\partial w} \right ]^{-1}$ in the optimal point, we have:
\begin{equation} 
\left [ \frac{\partial^2 \mathcal{L}_1}{\partial w\partial w} \right ]^{-1}=\gamma(I-I+\gamma \frac{\partial^2 \mathcal{L}_1}{\partial w\partial w})^{-1}=\gamma \sum_{j=0}^{\infty}\left [ I- \gamma \frac{\partial^2 \mathcal{L}_1}{\partial w\partial w} \right ]^j.
\end{equation}

So that:
\begin{equation} 
\nabla_{\alpha}\mathcal{L}_2=\frac{\partial \mathcal{L}_2}{\partial \alpha}-\gamma \frac{\partial \mathcal{L}_2}{\partial w}\sum_{j=0}^{\infty}\left [ I- \gamma \frac{\partial^2 \mathcal{L}_1}{\partial w\partial w} \right ]^j \frac{\partial^2 \mathcal{L}_1}{\partial \alpha \partial w}.
\end{equation}
\end{proof3}

\newenvironment{Theorem_1}{\paragraph{Proof of Theorem \ref{theorem1}}}{  $\square$}
\begin{Theorem_1}

Based on the Eq. \eqref{eq:neumann_approx_hypergradient} and \eqref{eq:darts_neuman}, we have
\begin{equation} 
\nabla_{\alpha}\mathcal{L}_2-\nabla_{\alpha}\tilde{\mathcal{L}}_2=\gamma \frac{\partial \mathcal{L}_2}{\partial w}\sum_{j=K+1}^{\infty}\left [ I- \gamma \frac{\partial^2 \mathcal{L}_1}{\partial w\partial w} \right ]^j \frac{\partial^2 \mathcal{L}_1}{\partial \alpha \partial w}.
\end{equation}

Since the $\mathcal{L}_1$ is $\mu$-strongly convex, and $\gamma \mu I\preceq \gamma \frac{\partial^2 \mathcal{L}_1}{\partial w\partial w}\preceq I$, we have 
\begin{equation} 
\sum_{j=K+1}^{\infty}\left [ I- \gamma \frac{\partial^2 \mathcal{L}_1}{\partial w\partial w} \right ]^j\leq \sum_{j=K+1}^{\infty} \left [ I- \gamma \mu \right ]^j.
\end{equation}
Based on the sum of geometric sequence, we have
\begin{equation} 
\sum_{j=K+1}^{\infty}\left [ I- \gamma \frac{\partial^2 \mathcal{L}_1}{\partial w\partial w} \right ]^j\leq \frac{1}{\gamma \mu}(1-\gamma \mu)^{K+1}.
\end{equation}

Since $\frac{\partial \mathcal{L}_2}{\partial w}$ and $\frac{\partial^2 \mathcal{L}_1}{\partial \alpha \partial w}$ are bounded, we have
\begin{equation} 
\left \|\nabla_{\alpha}\mathcal{L}_2 -\nabla_{\alpha}\tilde{\mathcal{L}}_2  \right \| \leqslant  C_{\mathcal{L}_1^{w\alpha}}\ C_{\mathcal{L}_2^w}\ \frac{1}{\mu }(1-\gamma \mu )^{K+1}.
\end{equation}


\end{Theorem_1}

\newenvironment{Corollary_1}{\paragraph{Proof: Corollary \ref{corollary2}}}{  $\square$}
\begin{Corollary_1}
Based on the definitions, the hypergradient of truncated back-propagation and the proposed Neumann approximation based hypergradient are defined in Eq.\eqref{eq:all_step} and Eq.\eqref{eq:neumann_approx_hypergradient}. When we assume that $w_t$ has converged to a stationary point $w^*$ in the last $K$ steps, we have 
\begin{equation} 
\begin{aligned}
w_{i}(\alpha)=w_{j}(\alpha)=w^*(\alpha), \qquad  for \ all\ i, j \in [T-K+1,T];\\
\frac{\partial \Phi(w_{i},\alpha)}{\partial w_{i}}=\frac{\partial \Phi(w_{j},\alpha)}{\partial w_{j}}=\frac{\partial \Phi(w^*(\alpha),\alpha)}{\partial w^*(\alpha)}=A_T, \qquad  for \ all\ i, j \in [T-K+1,T];\\
\frac{\partial \Phi(w_{i},\alpha)}{\partial \alpha}=\frac{\partial \Phi(w_{j},\alpha)}{\partial \alpha}=\frac{\partial \Phi(w^*(\alpha),\alpha)}{\partial \alpha}=B_T, \qquad  for \ all\ i, j \in [T-K+1,T].
\end{aligned}
\end{equation}

Now the truncated back-propagation could be formulated as:
\begin{equation} 
\begin{aligned}
h_{T-K}&=\frac{\partial \mathcal{L}_2}{\partial \alpha}+\frac{\partial \mathcal{L}_2}{\partial w_T}(\sum_{t=T-K+1}^{T}B_{t}A_{t+1}...A_T)\\
&=\frac{\partial \mathcal{L}_2}{\partial \alpha}+\frac{\partial \mathcal{L}_2}{\partial w_T}(\sum_{t=0}^{K}B_{T}A_T^t).\\
\end{aligned}
\end{equation}
We have 
\begin{equation} 
\begin{aligned}
A_T=\frac{\partial \Phi(w^*(\alpha),\alpha)}{\partial w^*(\alpha)}=\frac{\partial (w^*-\eta\frac{\partial \mathcal{L}_1}{\partial w})}{\partial w^*}=I-\gamma\frac{\partial^2 \mathcal{L}_1(w^*)}{\partial w\partial w},\\
B_T=\frac{\partial \Phi(w^*(\alpha),\alpha)}{\partial \alpha}=\frac{\partial (w^*-\eta\frac{\partial \mathcal{L}_1}{\partial w})}{\partial \alpha}=-\gamma\frac{\partial^2 \mathcal{L}_1(w^*)}{\partial \alpha \partial w}.\\
\end{aligned}
\end{equation}

From the above, we have
\begin{equation} 
\begin{aligned}
h_{T-K}&=\frac{\partial \mathcal{L}_2}{\partial \alpha}+\frac{\partial \mathcal{L}_2}{\partial w_T}(\sum_{t=0}^{K}B_{T}A_T^t)\\
&=\frac{\partial \mathcal{L}_2}{\partial \alpha}-\gamma \frac{\partial \mathcal{L}_2}{\partial w}\sum_{j=0}^{K}\left [ I- \gamma \frac{\partial^2 \mathcal{L}_1}{\partial w\partial w} \right ]^j \frac{\partial^2 \mathcal{L}_1}{\partial \alpha \partial w}\\
&=\nabla_{\alpha}\tilde{\mathcal{L}}_2.
\end{aligned}
\end{equation}


\end{Corollary_1}

\newenvironment{Lemma_4}{\paragraph{Proof of Lemma \ref{lemma_Lipschitz_differentiable}:}}{  $\square$}
\begin{Lemma_4}




First, for $\forall(\alpha, \alpha')$, we have
\begin{equation} \label{eq:27}
\begin{aligned}
&\left \|\nabla_{\alpha}\mathcal{L}_2(w,\alpha) -\nabla_{\alpha}\mathcal{L}_2(w,\alpha')  \right \|=\left \|\nabla_{\alpha}\mathcal{L}_2(\cdot,\alpha) -\nabla_{\alpha}\mathcal{L}_2(\cdot ,\alpha') + \nabla_{\alpha}\mathcal{L}_2(w(\alpha),\cdot) -\nabla_{\alpha}\mathcal{L}_2(w(\alpha'),\cdot) \right \|\\
=&\left \|\nabla_{\alpha}\mathcal{L}_2(\cdot,\alpha) -\nabla_{\alpha}\mathcal{L}_2(\cdot ,\alpha') + \nabla_{w}\mathcal{L}_2(w(\alpha),\cdot)\nabla_{\alpha}w(\alpha) -\nabla_{w}\mathcal{L}_2(w(\alpha'),\cdot)\nabla_{\alpha}w(\alpha') \right \|\\
\leq &\left \|\nabla_{\alpha}\mathcal{L}_2(\cdot,\alpha) -\nabla_{\alpha}\mathcal{L}_2(\cdot ,\alpha')\right \| + \left \|  \nabla_{w}\mathcal{L}_2(w(\alpha),\cdot)\nabla_{\alpha}w(\alpha) -\nabla_{w}\mathcal{L}_2(w(\alpha'),\cdot)\nabla_{\alpha}w(\alpha')\right \|. \\
\end{aligned}
\end{equation}
Then we divide Eq.\eqref{eq:27} to two parts. For the first part, based on the Assumption \ref{assumption1}.2, we have:
\begin{equation} \label{eq:28}
\begin{aligned}
\left \|\nabla_{\alpha}\mathcal{L}_2(\cdot,\alpha) -\nabla_{\alpha}\mathcal{L}_2(\cdot ,\alpha')\right \| \leq L^{\nabla_\alpha}_2 (\alpha -\alpha'). \\
\end{aligned}
\end{equation}
And for the second part of Eq.\eqref{eq:27}, we have
\begin{equation} \label{eq:29}
\resizebox{.9\linewidth}{!}{$
    \displaystyle
\begin{aligned}
&\left \|  \nabla_{w}\mathcal{L}_2(w(\alpha),\cdot)\nabla_{\alpha}w(\alpha) -\nabla_{w}\mathcal{L}_2(w(\alpha'),\cdot)\nabla_{\alpha}w(\alpha')\right \|\\
=&\left \|  \nabla_{w}\mathcal{L}_2(w(\alpha),\cdot)\nabla_{\alpha}w(\alpha) -\nabla_{w}\mathcal{L}_2(w(\alpha'),\cdot)\nabla_{\alpha}w(\alpha) -\nabla_{w}\mathcal{L}_2(w(\alpha'),\cdot)\nabla_{\alpha}w(\alpha')+\nabla_{w}\mathcal{L}_2(w(\alpha'),\cdot)\nabla_{\alpha}w(\alpha)\right \|\\
\leq&\left \|  \nabla_{w}\mathcal{L}_2(w(\alpha'),\cdot) -\nabla_{w}\mathcal{L}_2(w(\alpha'),\cdot)\right \|\left \| \nabla_{\alpha}w(\alpha) \right \| +\left \| \nabla_{w}\mathcal{L}_2(w(\alpha'),\cdot) \right \|\left \| \nabla_{\alpha}w(\alpha)-\nabla_{\alpha}w(\alpha') \right \|.
\end{aligned}
$}
\end{equation}

Based Assumption \ref{assumption1}.3, we have
\begin{equation} \label{eq:}
\left \|  \nabla_{w}\mathcal{L}_2(w(\alpha'),\cdot) -\nabla_{w}\mathcal{L}_2(w(\alpha'),\cdot)\right \|\leq L^{\nabla_w}_2 \left \| w(\alpha)-w(\alpha') \right \|, 
\end{equation}
and based Assumption \ref{assumption2}.2 that we have
\begin{equation} \label{eq:}
\left \| w(\alpha)-w(\alpha') \right \|\leq L_w \left \| \alpha- \alpha' \right \|,  \quad  \textup{and}\quad \left \| \nabla_{\alpha} w(\alpha)-\nabla_{\alpha}w(\alpha') \right \| \leq L_{\nabla_{\alpha}w} \left \| \alpha- \alpha' \right \|.
\end{equation}

Based on Assumption \ref{assumption1}.3, we know $\nabla_{w}\mathcal{L}_2(w(\alpha'),\cdot)$ is bounded that $\nabla_{w}\mathcal{L}_2(w(\alpha'),\cdot)\leq L^{w}_2$. $ \nabla_{\alpha}w(\alpha)$ is also bounded by $\left \| \nabla_{\alpha}w(\alpha)\right \| \leq L_w$. In this way, Eq.\eqref{eq:29} could be rephrased as:
\begin{equation} \label{eq:32}
\begin{aligned}
&\left \|  \nabla_{w}\mathcal{L}_2(w(\alpha),\cdot)\nabla_{\alpha}w(\alpha) -\nabla_{w}\mathcal{L}_2(w(\alpha'),\cdot)\nabla_{\alpha}w(\alpha')\right \| \leq L^{\nabla_w}_2 L_w^2 \left \| \alpha- \alpha' \right \|+ L^{w}_2 L_{\nabla_{\alpha}w} \left \| \alpha- \alpha' \right \|.
\end{aligned}
\end{equation}

Based on Eq. \eqref{eq:27}, Eq. \eqref{eq:28} and \eqref{eq:32} we have
\begin{equation} \label{eq:}
\left \|\nabla_{\alpha}\mathcal{L}_2(w,\alpha) -\nabla_{\alpha}\mathcal{L}_2(w,\alpha')  \right \|\leq (L^{\nabla_\alpha}_2 + L^{\nabla_w}_2 L_w^2 + L^{w}_2 L_{\nabla_{\alpha}w})\left \| \alpha- \alpha' \right \|.
\end{equation}

Therefore, Lemma 4 is proved.

\end{Lemma_4}

\newenvironment{Theorem_2}{\paragraph{Proof of Theorem \ref{theorem2}:}}{  $\square$}
\begin{Theorem_2}
We first define the noise term between the stochastic estimate $\nabla_{\alpha}\mathcal{L}^i_2$ and the true gradient $\nabla_{\alpha}\mathcal{L}_2$ as:
\begin{equation} \label{[t2]}
\varepsilon_i=\nabla_{\alpha}\mathcal{L}_2-\nabla_{\alpha}\mathcal{L}^i_2,
\end{equation}
and the error between the approximated hypergradient $\nabla_{\alpha}\tilde{\mathcal{L}}_2$ and the exact hypergradient $\nabla_{\alpha}\mathcal{L}_2$ as:
\begin{equation} \label{[t2]}
e_m=\nabla_{\alpha}\mathcal{L}_2(w^*(\alpha_m),\alpha_m)-\nabla_{\alpha}\tilde{\mathcal{L}}_2(w^*(\alpha_m),\alpha_m).
\end{equation}



We then prove that $\nabla_{\alpha}\mathcal{L}_2^i(w^*(\alpha_m),\alpha_m)$ is an unbiased estimate of $\nabla_{\alpha}\mathcal{L}_2(w^*(\alpha_m),\alpha_m)$ that:
\begin{equation} \label{eq:}
E[\nabla_{\alpha}\mathcal{L}_2^i(w^*(\alpha_m),\alpha_m)\mid \alpha_m]= \nabla_{\alpha}\mathcal{L}_2(w^*(\alpha_m),\alpha_m).
\end{equation}

Based on IFT in Eq.\eqref{eq:darts_neuman}, we have
\begin{equation} \label{eq:}
\resizebox{.9\linewidth}{!}{$
    \displaystyle
\begin{aligned}
&\nabla_{\alpha}\mathcal{L}_2^i(w^*(\alpha_m),\alpha_m)=\frac{\partial \mathcal{L}_2^i(w^*(\alpha_m),\alpha_m)}{\partial \alpha}-\frac{\partial\mathcal{L}_2^i(w^*(\alpha_m),\alpha_m)}{\partial w}\left [ \frac{\partial^2 \mathcal{L}_1^j(w^*(\alpha_m),\alpha_m)}{\partial w\partial w} \right ]^{-1}\frac{\partial^2 \mathcal{L}_1^j(w^*(\alpha_m),\alpha_m)}{\partial \alpha \partial w}.
\end{aligned}
$}
\end{equation}

So that 
\begin{equation} \label{eq:}
\begin{aligned}
&E\left[\nabla_{\alpha}\mathcal{L}_2^i(w^*(\alpha_m),\alpha_m) \mid \alpha_m\right ] \\
=&E\left [ \frac{\partial \mathcal{L}_2^i(w^*(\alpha_m),\alpha_m)}{\partial \alpha}-\frac{\partial\mathcal{L}_2^i(w^*(\alpha_m),\alpha_m)}{\partial w}\left [ \frac{\partial^2 \mathcal{L}_1^j(w^*(\alpha_m),\alpha_m)}{\partial w\partial w} \right ]^{-1}\frac{\partial^2 \mathcal{L}_1^j(w^*(\alpha_m),\alpha_m)}{\partial \alpha \partial w} \mid \alpha_m \right ].
\end{aligned}
\end{equation}

Based on the linear assumption for $\mathcal{L}_1^j$ in the condition 4 of the Theorem \ref{theorem2}, we have $ \frac{\partial^2 \mathcal{L}_1^j(w^*(\alpha_m),\alpha_m)}{\partial w\partial w} = \frac{\partial^2 \mathcal{L}_1(w^*(\alpha_m),\alpha_m)}{\partial w\partial w}$, and 
\begin{equation} \label{eq:}
\begin{aligned}
&E\left[\nabla_{\alpha}\mathcal{L}_2^i(w^*(\alpha_m),\alpha_m) \mid \alpha_m\right ]\\ &=\frac{1}{R}\sum_{i=1}^{R}\frac{\partial \mathcal{L}_2^i(w^*(\alpha_m),\alpha_m)}{\partial \alpha}-\frac{1}{R}\sum_{i=1}^{R}\frac{\partial\mathcal{L}_2^i(w^*(\alpha_m),\alpha_m)}{\partial w}\left [ \frac{\partial^2 \mathcal{L}_1(w^*(\alpha_m),\alpha_m)}{\partial w\partial w} \right ]^{-1}\frac{1}{J}\sum_{j=1}^{J}\frac{\partial^2 \mathcal{L}_1^j(w^*(\alpha_m),\alpha_m)}{\partial \alpha \partial w}\\
&=\frac{\partial \mathcal{L}_2(w^*(\alpha_m),\alpha_m)}{\partial \alpha}-\frac{\partial\mathcal{L}_2(w^*(\alpha_m),\alpha_m)}{\partial w}\left [ \frac{\partial^2 \mathcal{L}_1(w^*(\alpha_m),\alpha_m)}{\partial w\partial w} \right ]^{-1}\frac{\partial^2 \mathcal{L}_1(w^*(\alpha_m),\alpha_m)}{\partial \alpha \partial w}\\
&=\nabla_{\alpha}\mathcal{L}_2(w^*(\alpha_m),\alpha_m).
\end{aligned}
\end{equation}

Based on the Lemma \ref{lemma_Lipschitz_differentiable}, we know that $\nabla_{\alpha}\mathcal{L}_2(w^*(\alpha_m),\alpha_m)$ is Lipschitz continuous with $L_{\nabla_\alpha \mathcal{L}_2}=L^{\nabla_\alpha}_2 + L^{\nabla_w}_2 L_w^2 + L^{w}_2 L_{\nabla_{\alpha}w}$. Based on Lipschitz condition, we have 
\begin{equation} \label{eq:}
\begin{aligned}
&E\left[\mathcal{L}_2(w^*(\alpha_{m+1}),\alpha_{m+1}) \mid \alpha_m\right ] \leq E\left[\mathcal{L}_2(w^*(\alpha_{m}),\alpha_{m}) \mid \alpha_m\right ]\\
&+E\left[\left \langle \nabla_{\alpha}\mathcal{L}_2(w^*(\alpha_{m}),\alpha_{m}),\alpha_{m+1}-\alpha_m \right \rangle \mid \alpha_m\right ]+\frac{L_{\nabla_\alpha \mathcal{L}_2}}{2}E\left [ \left \| \alpha_{m+1}-\alpha_{m} \right \|^2 \right ]\\ 
& =\mathcal{L}_2(w^*(\alpha_{m}),\alpha_{m})+\left \langle E\left[\nabla_{\alpha}\mathcal{L}_2(w^*(\alpha_{m}),\alpha_{m})\right ],-\gamma_{\alpha_m} E\left [ \nabla_{\alpha}\mathcal{L}_2^{i'}(w^*(\alpha_{m}),\alpha_{m})   \mid \alpha_m\right ]\right \rangle\\
&+\frac{L_{\nabla_\alpha \mathcal{L}_2}}{2} \gamma_{\alpha_m}^2 E\left [ \left \| \nabla_{\alpha}\mathcal{L}_2^{i'}(w^*(\alpha_{m}),\alpha_{m}) \right \| ^2 \right ].\\
\end{aligned}
\end{equation}

From our definitions, we have
\begin{equation} \label{eq:}
\resizebox{.9\linewidth}{!}{$
    \displaystyle
\begin{aligned}
E\left[\nabla_{\alpha}\mathcal{L}_2(w^*(\alpha_{m}),\alpha_{m})\right ]&=E\left[\nabla_{\alpha}\tilde{\mathcal{L}}_2(w^*(\alpha_{m}),\alpha_{m})+e_m\right ]=E\left[\nabla_{\alpha}\tilde{\mathcal{L}}_2(w^*(\alpha_{m}),\alpha_{m})\right ]+E\left [ e_m  \right ],\\
E\left [ \nabla_{\alpha}\hat{\mathcal{L}}_2^{i}(w^*(\alpha_{m}),\alpha_{m}) \mid \alpha_m \right ]&=E\left [ \nabla_{\alpha}\tilde{\mathcal{L}}_2(w^*(\alpha_{m}),\alpha_{m}) -\varepsilon_m \mid \alpha_m \right ]=E\left [ \nabla_{\alpha}\tilde{\mathcal{L}}_2(w^*(\alpha_{m}),\alpha_{m}) \right ],  \\
E\left [ \left \| \nabla_{\alpha}\hat{\mathcal{L}}_2^i(w^j(\alpha_m),\alpha_m) \right \| ^2\mid \alpha_m \right ]&=E\left [ \left \| \nabla_{\alpha}\tilde{\mathcal{L}}_2(w^*(\alpha_{m}),\alpha_{m})-\varepsilon_m  \right \| ^2 \right ]= E\left [ \left \| \nabla_{\alpha}\tilde{\mathcal{L}}_2(w^*(\alpha_{m}),\alpha_{m})  \right \| ^2 \right ]+E\left [ \left \| \varepsilon_m  \right \| ^2 \right ],
\end{aligned}
$}
\end{equation}

since $E(\varepsilon_m)=0$. In this way, we have
\begin{equation} \label{eq:}
\begin{aligned}
E\left[\mathcal{L}_2(w^*(\alpha_{m+1}),\alpha_{m+1}) \mid \alpha_m\right ] & \leq E\left[\mathcal{L}_2(w^*(\alpha_{m}),\alpha_{m}) \mid \alpha_m\right ]-\gamma_{\alpha_m} E\left [ \left \| \nabla_{\alpha}\tilde{\mathcal{L}}_2(w^*(\alpha_{m}),\alpha_{m})  \right \| ^2 \right ]\\
&-\gamma_{\alpha_m} E \left \langle  e_m, \nabla_{\alpha}\tilde{\mathcal{L}}_2(w^*(\alpha_{m}),\alpha_{m}) \right \rangle +\frac{L_{\nabla_\alpha \mathcal{L}_2}}{2}\gamma_{\alpha_m}^2E\left [ \left \| \nabla_{\alpha}\tilde{\mathcal{L}}_2(w^*(\alpha_{m}),\alpha_{m})  \right \| ^2 \right ]\\
&+\frac{L_{\nabla_\alpha \mathcal{L}_2}}{2}\gamma_{\alpha_m}^2 E\left [ \left \| \varepsilon_m  \right \| ^2 \right ].\\ 
\end{aligned}
\end{equation}

Based on Theorem \ref{theorem1}, we have $\left \| e_m \right \| \leqslant  C_{\mathcal{L}_1^{w\alpha}}\ C_{\mathcal{L}_2^w}\ \frac{1}{\mu }(1-\gamma \mu )^{K+1}$. In this way, for all $\nabla_{\alpha}\tilde{\mathcal{L}}_2(w^*(\alpha_{m}),\alpha_{m})$, we have

\begin{equation} \label{eq:}
\begin{aligned}
\left \langle e_m, \nabla_{\alpha}\tilde{\mathcal{L}}_2(w^*(\alpha_{m}),\alpha_{m}) \right \rangle &\geq -C_{\mathcal{L}_1^{w\alpha}}\ C_{\mathcal{L}_2^w}\ \frac{1}{\mu }(1-\gamma \mu )^{K+1} \left \|  \nabla_{\alpha}\tilde{\mathcal{L}}_2 \right \| \\
&=-\frac{C_{\mathcal{L}_1^{w\alpha}}\ C_{\mathcal{L}_2^w}\ (1-\gamma \mu )^{K+1}}{\mu \left \|  \nabla_{\alpha}\tilde{\mathcal{L}}_2 \right \| } \left \|  \nabla_{\alpha}\tilde{\mathcal{L}}_2 \right \|^2\\
&=-P \left \|  \nabla_{\alpha}\tilde{\mathcal{L}}_2 \right \|^2,
\end{aligned}
\end{equation}
where $P=\frac{C_{\mathcal{L}_1^{w\alpha}}\ C_{\mathcal{L}_2^w}\ (1-\gamma \mu )^{K+1}}{\mu \left \|  \nabla_{\alpha}\tilde{\mathcal{L}}_2 \right \|}$. In this way, we have:
\begin{equation} \label{eq:}
\resizebox{.85\linewidth}{!}{$
    \displaystyle
\begin{aligned}
E\left[\mathcal{L}_2(w^*(\alpha_{m+1}),\alpha_{m+1}) \right ] & \leq E\left[\mathcal{L}_2(w^*(\alpha_{m}),\alpha_{m}) \right ]-\gamma_{\alpha_m}(1-P) E\left [ \left \| \nabla_{\alpha}\tilde{\mathcal{L}}_2  \right \| ^2 \right ]\\
&+\frac{L_{\nabla_\alpha \mathcal{L}_2}}{2}\gamma_{\alpha_m}^2(1+D)E\left [ \left \| \nabla_{\alpha}\tilde{\mathcal{L}}_2 \right \| ^2 \right ]\\ 
&\leq E\left[\mathcal{L}_2(w^*(\alpha_{m}),\alpha_{m}) \right ]-\gamma_{\alpha_m}[(1-P)-\frac{L_{\nabla_\alpha \mathcal{L}_2}}{2}\gamma_{\alpha_m}(1+D)]E\left [ \left \| \nabla_{\alpha}\tilde{\mathcal{L}}_2 \right \| ^2 \right ].\\ 
\end{aligned}
$}
\end{equation}

If we choose $\gamma_{\alpha_m}$ to make $(1-P)-\frac{L_{\nabla_\alpha \mathcal{L}_2}}{2}\gamma_{\alpha_m}(1+D)>0$, we have $\gamma_{\alpha_m}<\frac{(1-P)}{\frac{L_{\nabla_\alpha \mathcal{L}_2}}{2}(1+D)}$. In addition, since the learning rate should be positive, we should make that $1-P>0$, which could be reached by choose appropriate $\gamma$ and $K$ that $\frac{C_{\mathcal{L}_1^{w\alpha}}\ C_{\mathcal{L}_2^w}\ (1-\gamma \mu )^{K+1}}{\mu \left \|  \nabla_{\alpha}\tilde{\mathcal{L}}_2 \right \| }<1$, where $0< 1-\gamma \mu \leq 1$. In this way, we could find that $\mathcal{L}_2$ is decreasing with $\alpha_m$, and we know that with sufficiently large $m$, $\mathcal{L}_2$ will decrease and converge  since $\mathcal{L}_2$ is bounded.

Furthermore, we have:
\begin{equation} \label{eq:}
\begin{aligned}
& E\left[\mathcal{L}_2(w^*(\alpha_{m}),\alpha_{m}) \right ]-E\left[\mathcal{L}_2(w^*(\alpha_{m+1}),\alpha_{m+1}) \right ] \\
\geq & \gamma_{\alpha_m}[(1-P)-\frac{L_{\nabla_\alpha \mathcal{L}_2}}{2}\gamma_{\alpha_m}(1+D)]E\left [ \left \| \nabla_{\alpha}\tilde{\mathcal{L}}_2(w^*(\alpha_{m}),\alpha_{m})  \right \| ^2 \right ].\\ 
\end{aligned}
\end{equation}

By telescoping sum, we can show that
\begin{equation} \label{eq:}
\begin{aligned}
E\left[\mathcal{L}_2(w^*(\alpha_{0}),\alpha_{0}) \right ]-E\left[\mathcal{L}_2(w^*(\alpha_{m}),\alpha_{m}) \right ] & \geq \sum_{k=0}^{K} q_t E\left [ \left \| \nabla_{\alpha}\tilde{\mathcal{L}}_2(w^*(\alpha_{m}),\alpha_{m})  \right \| ^2 \right ],\\ 
\end{aligned}
\end{equation}
where $q_t=\gamma_{\alpha_m}[(1-P)-\frac{L_{\nabla_\alpha \mathcal{L}_2}}{2}\gamma_{\alpha_m}(1+D)]>0$. Since $\mathcal{L}_2$ is bounded, we have $\underset{K\rightarrow \infty}{\sum_{k=0}^{K}} q_t E\left [ \left \| \nabla_{\alpha}\tilde{\mathcal{L}}_2(w^*(\alpha_{m}),\alpha_{m})  \right \| ^2 \right ]<\infty$. In addition, based on condition 3, we have $\underset{K\rightarrow \infty}{\sum_{k=0}^{K}} q_t=\infty$, which imply that $\underset{k\rightarrow \infty}{\textup{lim}}E\left [ \left \| \nabla_{\alpha}\tilde{\mathcal{L}}_2(w^*(\alpha_{m}),\alpha_{m})  \right \|  \right ]=0$, so as $\underset{m\rightarrow \infty}{\textup{lim}} E\left [ \left \| \nabla_{\alpha}\hat{\mathcal{L}}_2^{i}(w^j(\alpha_m),\alpha_m) \right \| \right ]=0$.

\end{Theorem_2}


\section*{B. Practical implementation of hypergradient}
\label{imple_hypergra}
As described, our iDARTS is built based on the DARTS framework with reformulation the hypergradient calculation as:
\begin{equation} \label{eq:}
\nabla_{\alpha}\tilde{\mathcal{L}}_2=\frac{\partial \mathcal{L}_2}{\partial \alpha}-\gamma \frac{\partial \mathcal{L}_2}{\partial w}\sum_{k=0}^{K}\left [ I- \gamma \frac{\partial^2 \mathcal{L}_1}{\partial w\partial w} \right ]^k \frac{\partial^2 \mathcal{L}_1}{\partial \alpha \partial w}.
\end{equation}
where the different part is the $\left [ I- \gamma \frac{\partial^2 \mathcal{L}_1}{\partial w\partial w} \right ]^k$. As known, it is costly to calculate the Hessian matrix $\frac{\partial^2 \mathcal{L}_1}{\partial w\partial w}$ and $\frac{\partial^2 \mathcal{L}_1}{\partial \alpha \partial w}$ for a large neural network, and we propose two approximations to reduce the computational cost for pracyical implementation, as described below.

\textbf{Approximation 1:} Although it is hard to directly calculate  the Hessian matrix $\frac{\partial^2 \mathcal{L}_1}{\partial w\partial w}$, we could consider Hessian-vector product technique with autograd to calculate $\frac{\partial \mathcal{L}_2}{\partial w} \cdot \frac{\partial^2 \mathcal{L}_1}{\partial w\partial w}$. In this way, we can calculate $\frac{\partial \mathcal{L}_2}{\partial w}\sum_{k=0}^{K}\left [ I- \gamma \frac{\partial^2 \mathcal{L}_1}{\partial w\partial w} \right ]^k$ step by step:
\begin{equation} \label{eq:}
\begin{aligned}
\frac{\partial \mathcal{L}_2}{\partial w}\sum_{k=0}^{K}\left [ I- \gamma \frac{\partial^2 \mathcal{L}_1}{\partial w\partial w} \right ]^k =\sum_{k=0}^{K} \frac{\partial \mathcal{L}_2}{\partial w}\left[ I- \gamma \frac{\partial^2 \mathcal{L}_1}{\partial w\partial w} \right ]^k=\sum_{k=0}^{K} V_0\left[ I- \gamma H \right ]^k=V_0+V_1+V_2+...+V_k.
\end{aligned}
\end{equation}
where we define $V_0=\frac{\partial \mathcal{L}_2}{\partial w}$, $H=\gamma \frac{\partial^2 \mathcal{L}_1}{\partial w\partial w}$, and $V_1=V_0(I-H), V_2=V_1(I-H),...,V_K=V_{K-1}(I-H)$. We can find that, $\frac{\partial \mathcal{L}_2}{\partial w}\sum_{k=0}^{K}\left [ I- \gamma \frac{\partial^2 \mathcal{L}_1}{\partial w\partial w} \right ]^k$ could be calculated with $K$ steps of Hessian-vector product. 




\textbf{Approximation 2:} Apart from the Hessian matrix $\frac{\partial^2 \mathcal{L}_1}{\partial w\partial w}$, it is also costly to calculate $\frac{\partial^2 \mathcal{L}_1}{\partial \alpha \partial w}$ for large neural networks, and we follow DARTS to use the Taylor expansion to approximate $\frac{\partial^2 \mathcal{L}_1}{\partial \alpha \partial w}$. After calculating $\frac{\partial \mathcal{L}_2}{\partial w}\sum_{k=0}^{K}\left [ I- \gamma \frac{\partial^2 \mathcal{L}_1}{\partial w\partial w} \right ]^k$, considering the function $\frac{\partial \mathcal{L}_1(w, \alpha)}{\partial \alpha}$ with Taylor expansion, we have
\begin{equation} \label{eq:}
\begin{aligned}
\frac{\partial \mathcal{L}_1(w+\epsilon A, \alpha)}{\partial \alpha}=\frac{\partial \mathcal{L}_1(w, \alpha)}{\partial \alpha}+\frac{\partial^2 \mathcal{L}_1(, \alpha)}{\partial \alpha \partial w}\epsilon A +...,\\
\frac{\partial \mathcal{L}_1(w-\epsilon A, \alpha)}{\partial \alpha}=\frac{\partial \mathcal{L}_1(w, \alpha)}{\partial \alpha}-\frac{\partial^2 \mathcal{L}_1(w, \alpha)}{\partial \alpha \partial w}\epsilon A +...,\\
\end{aligned}
\end{equation}
where $\epsilon$ is a very small scalar. When we replace $A$ with $\frac{\partial \mathcal{L}_2}{\partial w}\sum_{k=0}^{K}\left [ I- \gamma \frac{\partial^2 \mathcal{L}_1}{\partial w\partial w} \right ]^k$, we have
\begin{equation} \label{eq:}
\frac{\partial \mathcal{L}_2}{\partial w}\sum_{k=0}^{K}\left [ I- \gamma \frac{\partial^2 \mathcal{L}_1}{\partial w\partial w} \right ]^k \frac{\partial^2 \mathcal{L}_1}{\partial \alpha \partial w}= \frac{\frac{\partial \mathcal{L}_1(w+\epsilon A, \alpha)}{\partial \alpha}-\frac{\partial \mathcal{L}_1(w-\epsilon A, \alpha)}{\partial \alpha}}{2\epsilon}.
\end{equation}

As described, the proposed approximated hypergradient $\nabla_{\alpha}\tilde{\mathcal{L}}_2$ is easy to implement based on the DARTS framework with only replacing $\frac{\partial \mathcal{L}_2}{\partial w}$ to $\frac{\partial \mathcal{L}_2}{\partial w}\sum_{k=0}^{K}\left [ I- \gamma \frac{\partial^2 \mathcal{L}_1}{\partial w\partial w} \right ]^k$, which could be computed using the the Hessian-vector product technique.

Therefor, we can practically implement our approximated hypergradient $\nabla_{\alpha}\tilde{\mathcal{L}}_2$, so as the stochastic approximated hypergradient $\nabla_{\alpha}\hat{\mathcal{L}}_2^i(w^j(\alpha),\alpha)$ with minibatches based on the DARTS framework \footnote{The codes and training log files could be found in the supplementary material. The best trained models on CIFAR-10, CIFR-100, and ImageNet could be found \href{https://github.com/MiaoZhang0525/iDARTS}{https://github.com/MiaoZhang0525/iDARTS}.}.




\begin{figure}
\centering
 \subfloat[Validation error with $T$=1]{
  \begin{minipage}{4cm}
      \includegraphics[width=4cm,height=3cm]{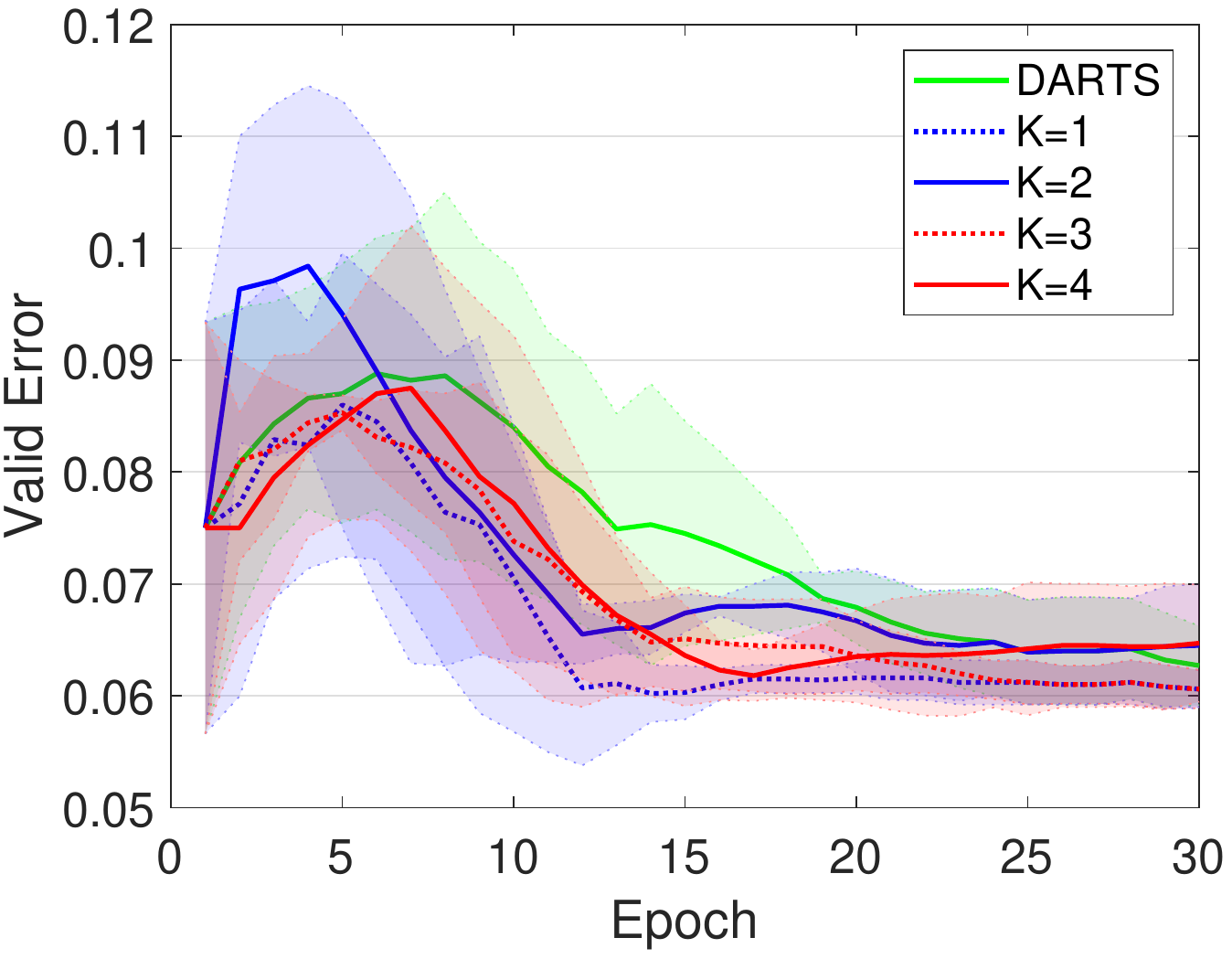}
  \end{minipage}
 }
  \subfloat[Test errors with $T$=1]{
  \begin{minipage}{4cm}
       \includegraphics[width=4cm,height=3cm]{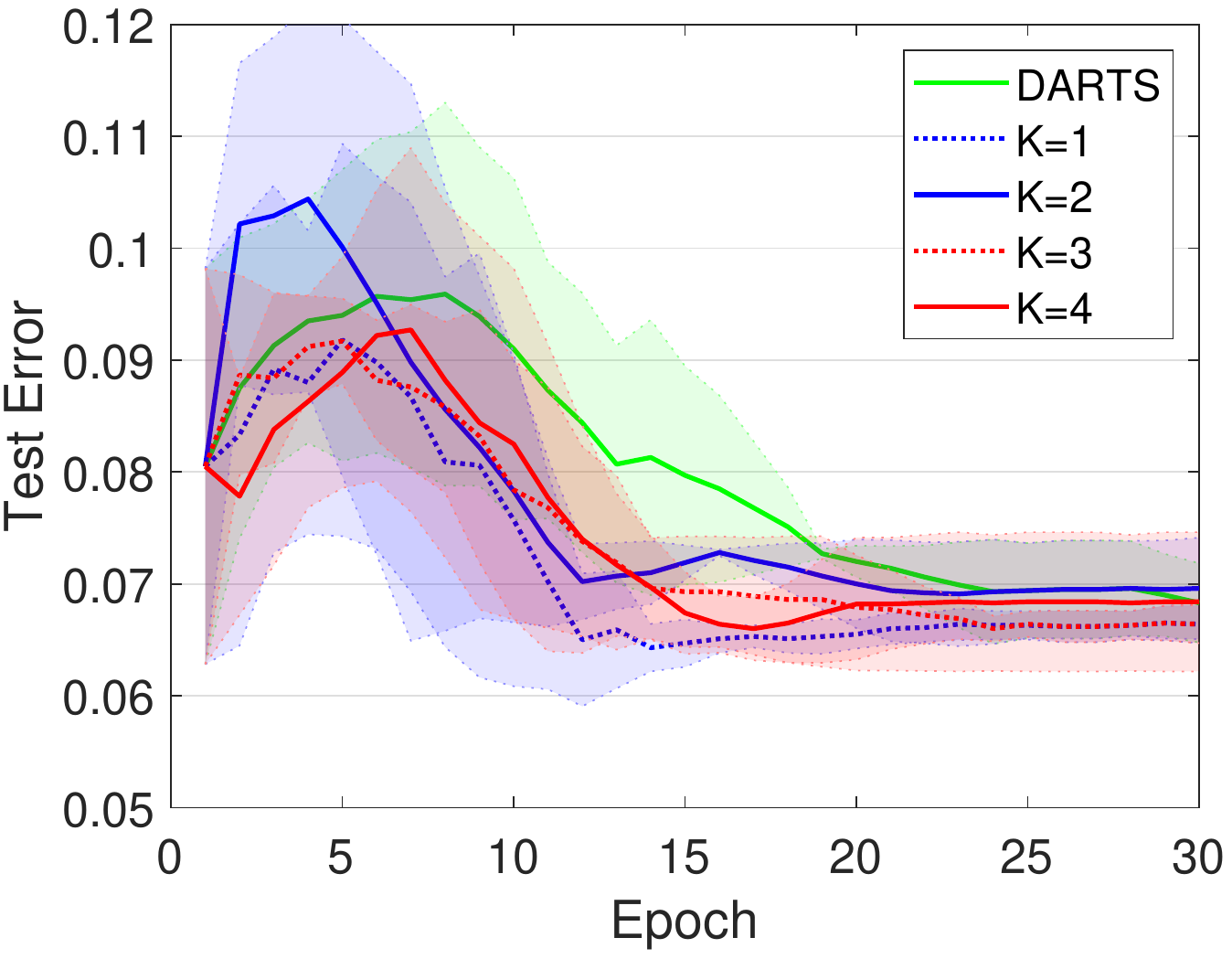}
  \end{minipage} 
  }
 \subfloat[Validation error with $T$=5]{
  \begin{minipage}{4cm}
      \includegraphics[width=4cm,height=3cm]{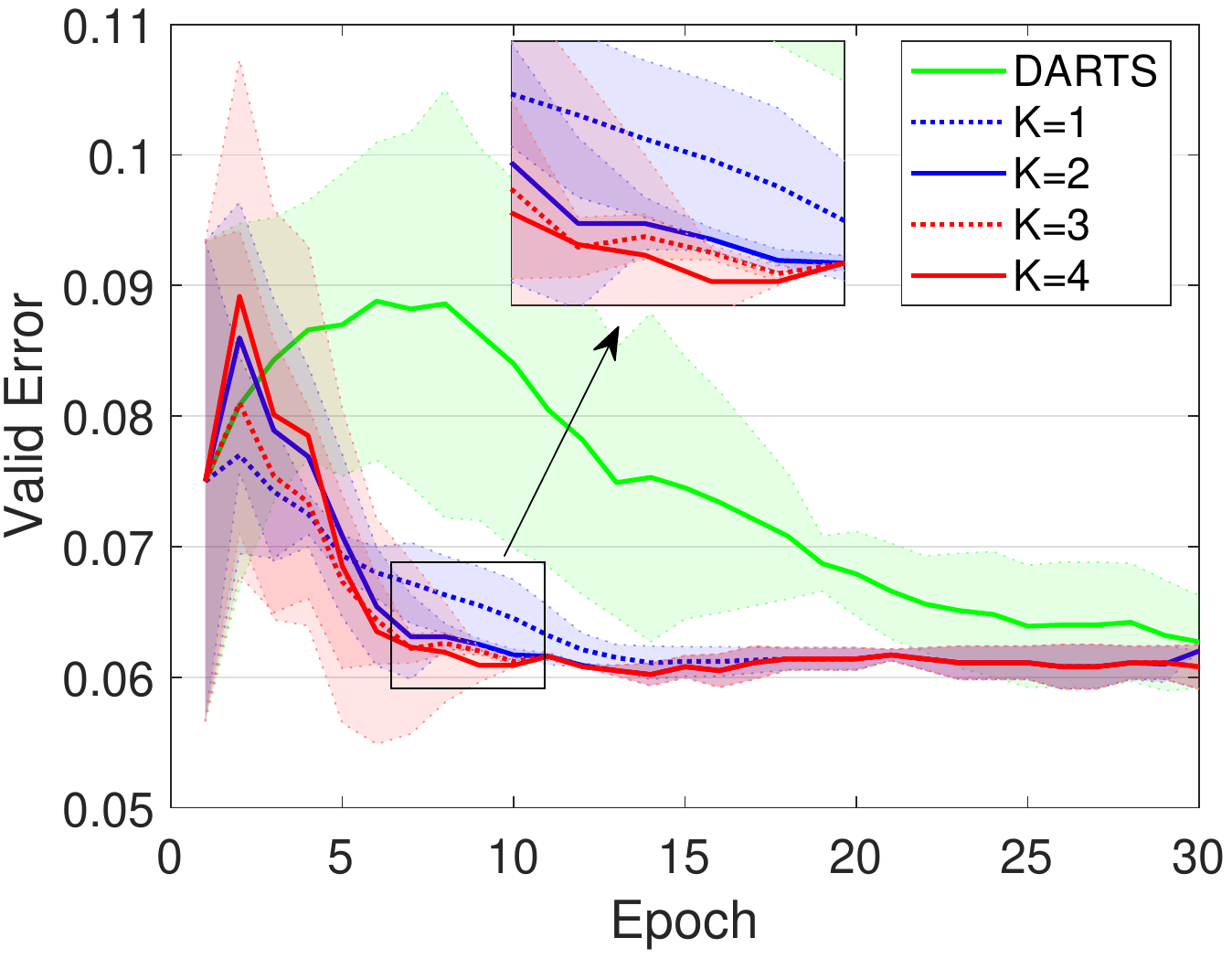}
  \end{minipage}
 }
  \subfloat[Test errors with $T$=5]{
  \begin{minipage}{4cm}
       \includegraphics[width=4cm,height=3cm]{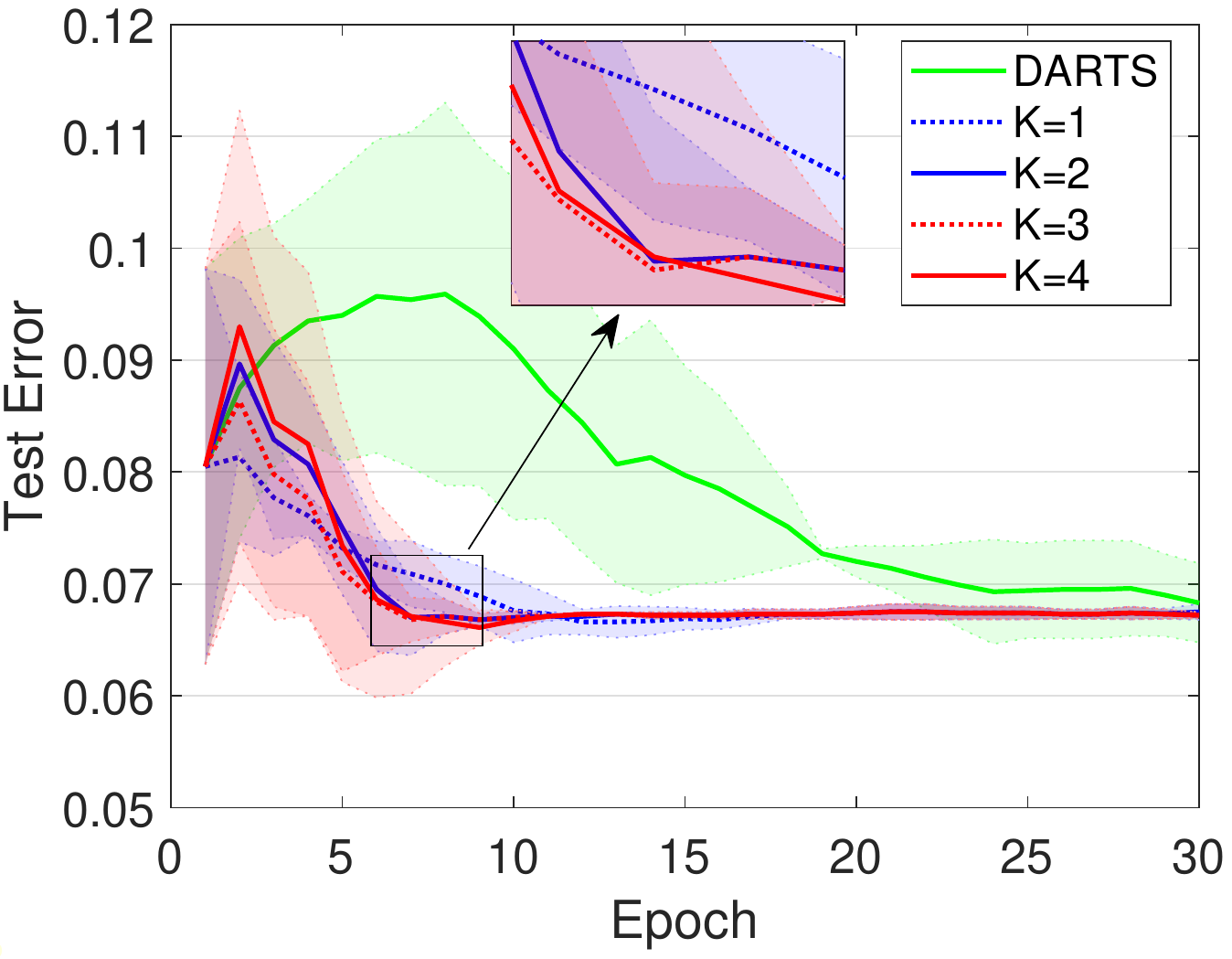}
  \end{minipage} 
  }  
 \caption{Ablation study on $K$ for iDARTS with $T=1$ and $T=5$ on NAS-Bench-1Shot1.}
 \label{fig:K_nas1shot1}
\end{figure}

\section*{C. Ablation study on the number of approximation terms $K$}
\label{app_K}
As we described before, there are two additional hyperparameters in our practical iDARTS, the inner optimization steps $T$ and the number of terms for the approximation in Eq.\eqref{eq:neumann_approx_hypergradient}. We have analyzed $T$ in previous experiments. In this section, we analyze another hyperparameter $K$ on the NAS-Bench-1Shot1 benchmark dataset. In the first experiment, we set a default hyperparameter $T=1$ the same as DARTS for the inner supernet training to remove the bias from $T$. From Eq.\eqref{eq:neumann_approx_hypergradient} and \eqref{eq:darts_hypergradient}, we could further find that the hypergradient calculation in our iDARTS with $T = 1$ and $K = 0$ is the same as DARTS. Figure \ref{fig:K_nas1shot1} (a) (b) plots the performance of iDARTS with different $K$ on the NAS-Bench-1Shot1. As shown, our iDARTS is very robust to $K$ with limited training steps $T=1$, where iDARTS with different $K$ all outperform the DARTS baseline with the same inner training steps $T=1$, showing the superiority of the proposed approximation over DARTS. Another interesting finding is that, our iDARTS with $K=1$ and $T=1$ even achieve slightly more competitive results than $K>1$. An underlying reason is that, when the inner training step is too small, it is hard to achieve the local optimal $w^*$ and the corresponding hypergradient is not accurate.

To further investigate the effectiveness of the proposed approximation, we consider setting enough inner training steps with $T=5$, and Figure \ref{fig:K_nas1shot1} (c) (d) plots the performance of iDARTS with different $K$ on the NAS-Bench-1Shot1 under $T=5$. The first impression from Figure \ref{fig:K_nas1shot1} is that increasing inner training steps could significantly improve the performance, where all cases with $T=5$ generally outperform $T=1$. Another interesting finding is that, with enough inner training steps, the number of approximation terms $K$ has a positive impact on the performance of iDARTS. As shown in Figure \ref{fig:K_nas1shot1} (c) (d), increasing $K$ also helps iDARTS converge to excellent solutions faster, verifying that the proposed $\nabla_{\alpha}\hat{\mathcal{L}}_2^i$ could asymptotically approach to the exact hypergradient $\nabla_{\alpha}\mathcal{L}_2^i$ with the increase of approximation term $K$. Besides, we can find that, $K=2$ is large enough to result in competitive performance for our iDARTS on NAS-Bench-1shot1, which results in similar performance as $K\geq 3$.

We also conduct an ablation study on NAS-Bench-201 dataset to analyse the hyperparameter $K$, and Table \ref{tab:nasbench201_K} summarizes the performance of iDARTS on NAS-Bench-201 with a different number of approximation term $K$. The results in Table \ref{tab:nasbench201_K} are similar to those on the NAS-Bench-1Shot1 dataset, also showing that $K$ has a positive impact on the performance of iDARTS. Firstly, we can find that, with the same inner training steps $T=1$ as DARTS baseline, our iDARTS ($T=1$, $K=1$) with one approximation term outperform DARTS by large margins in this case, verifying the superiority of the proposed approximation over DARTS. Secondly, the results in Table \ref{tab:nasbench201_K} also demonstrate that considering more approximation terms does indeed help improve our iDARTS to a certain degree. With enough inner training steps, the performance of iDARTS increases with $K$ from 0 to 2. Another interesting finding is that the performance of iDARTS does not always increase with the $K$, and there is a decrease for $K\geq 3$. One underlying reason may be that, the iDARTS with smaller $K$ brings more noises into the hypergradient, which in turn enhances the exploration. Several recent works \cite{chen2020stabilizing,miaozhang20} show the importance of the exploration in the differentiable NAS, where adding more noises into the hypergradient could improve the performance. Our experimental results suggest that a $K=2$ achieves an excellent trade-off between the accuracy of hypergradient and the exploration, thus achieving the competitive performance on the NAS-Bench-201 dataset.

\begin{table*}
\footnotesize
\centering
\caption{Ablation study on $K$ for iDARTS with on NAS-Bench-201. 
}
\begin{tabular}
{|l|c|c|c|c|c|c|c|c|}
\hline

\makecell[c]{\multirow{2}*{Method}}&\multicolumn{2}{c|}{CIFAR-10}&\multicolumn{2}{c|}{CIFAR-100}&\multicolumn{2}{c|}{ImageNet-16-120}\\
~&\multicolumn{1}{c}{Valid(\%)}&\multicolumn{1}{c|}{Test(\%)}&\multicolumn{1}{c}{Valid(\%)}&\multicolumn{1}{c|}{Test(\%) }&\multicolumn{1}{c}{Valid(\%) }&\multicolumn{1}{c|}{Test(\%)}\\
\hline
\hline
DARTS($T=1$, $K=0$)
&39.77$\pm$0.00&54.30$\pm$0.00&15.03$\pm$0.00&15.61$\pm$0.00&16.43$\pm$0.00&16.32$\pm$0.00\\
iDARTS($T=1$, $K=1$)
&86.85$\pm$0.93&89.67$\pm$1.31&64.09$\pm$2.92&64.17$\pm$3.26&36.26$\pm$5.71&36.11$\pm$ 5.77\\
iDARTS($T=4$, $K=0$) &87.31$\pm$1.33&90.36$\pm$1.79&64.76$\pm$2.54&64.43$\pm$2.47&32.53$\pm$1.31&32.42$\pm$1.54\\
iDARTS($T=4$, $K=1$) &89.30$\pm$1.47&92.44$\pm$1.14&67.88$\pm$1.86&68.17$\pm$2.81&37.11$\pm$7.79&36.61$\pm$7.47
\\
iDARTS($T=4$, $K=2$) &89.86$\pm$0.60&93.58$\pm$0.32&70.57$\pm$0.24&70.83$\pm$0.48&40.38$\pm$0.59&40.89$\pm$0.68\\
iDARTS($T=4$, $K=3$) &89.35$\pm$0.03&92.29$\pm$0.26&68.51$\pm$0.77& 68.58$\pm$1.18&42.37$\pm$0.48&42.26$\pm$0.41\\
\hline
\end{tabular}
\label{tab:nasbench201_K}
\end{table*}

\section*{D. Experimental settings in all experiments}

In the first experimental set, we choose the third search space of NAS-Bench-1Shot1 \cite{zela2020nasbench1shot1} to analyze iDARTS, since it is much more complicated than the remaining two search spaces and is a better case to identify the advantages of iDARTS. In Section \ref{sec5.1}, we analyzed the hyperparameter $T$ for our iDARTS and compared it with baseline on the NAS-Bench-1Shot1, and we set another hyperparameter $K=3$ in all cases. In Appendix C, we further conduct the ablation study to investigate another important hyperparameter $K$, where we consider two cases with $T=1$ and $T=5$, and the remaining experimental settings are the same as the default settings. 

In the second experimental set, we choose the NAS-Bench-201 dataset \cite{BENCH102} to analyze differentiable NAS methods. In Section \ref{sec5.2}, we first conduct a comparison experiment with several NAS baselines, and the hyperparameters for our iDARTS in this experiments are $T=4$, $K=2$, and $\gamma$=0.01. Then we conduct a series ablation studies to investigate three important hyperparameters, inner supernet training steps $T$, supernet learning rate $\gamma$, and architecture learning rate $\gamma_{\alpha}$. In the experiment for the investigation of $T$, see Figure \ref{fig:T_nas201} (a), we set $K=1$ and other hyperparameters are default settings. In the Figure \ref{fig:T_nas201} (b) and (c), we set $T=4$ and $K=1$ to investigate both the supernet learning rate $\gamma$ and architecture learning rate $\gamma_{\alpha}$. In Appendix C, we also analyze the impact of $K$ in iDARTS on NAS-Bench-201 dataset, where we set $T=4$ and $\gamma$=0.01, and the remaining settings are the default.

In the common DARTS search space, we follow the experimental settings in \cite{liu2018darts} to compare with the state-of-the-art NAS methods. We search for micro-cell structures on CIFAR-10 to stack more cells to form the final structure for architecture evaluation. There are two types of cells with the unified search space: a normal cell $\alpha_{normal}$ and a reduction cell $\alpha_{reduce}$. Cell structures are repeatedly stacked to form the final CNN structure. There are only two reduction cells in the final CNN structure, located in the 1/3 and 2/3 depths of the network. The best architecture searched by our iDARTS on the DARTS search space is obtained with $T=4$ and $K=2$. In CIFAR-10, we stack 20 cells to form the final structure for training. The batch size is set as 96, and the number of initial filters is 36. We then transfer the best-searched cells to CIFAR-100 and ImageNet to evaluate the transferability. The experiment setting for the evaluation in CIFAR-100 is the same as CIFAR-10. In the ImageNet dataset, the experiment setting is slightly different from CIFAR-10 in that only 14 cells are stacked, and the number of initial channels is changed to 48, and the batch size is set as 128. We use a linear learning rate scheduler and also following PDART \cite{chen2019progressive} and PCDARTS \cite{xu2019pcdarts} to use a smaller slope in the last five epochs for the architecture evaluation on the ImageNet.


\begin{table*}
\centering
\caption{An overview of different hypergradient approximations.}

\begin{tabular}{|l|c|c|c|}
\hline

\multicolumn{1}{|c|}{Method}&{Steps}&{Memory Cost}&{Hypergradient Calculation}\\
\hline\hline
Exact IFT hypergradient&$\infty$&$\mathcal{O}(P+H)$&$\frac{\partial \mathcal{L}_2}{\partial \alpha}-\frac{\partial \mathcal{L}_2}{\partial w}\left [ \frac{\partial^2 \mathcal{L}_1}{\partial w\partial w} \right ]^{-1}\frac{\partial^2 \mathcal{L}_1}{\partial \alpha \partial w}$\\
DARTS \cite{liu2018darts} &1&$\mathcal{O}(P+H)$&$\frac{\partial \mathcal{L}_2}{\partial \alpha}-\gamma \frac{\partial \mathcal{L}_2}{\partial w}\frac{\partial^2 \mathcal{L}_1}{\partial \alpha \partial w}$\\
$T_1-T_2$ \cite{luketina2016scalable} &1&$\mathcal{O}(P+H)$&$\frac{\partial \mathcal{L}_2}{\partial \alpha}-\frac{\partial \mathcal{L}_2}{\partial w} [I]^{-1} \frac{\partial^2 \mathcal{L}_1}{\partial \alpha \partial w}$\\
Reverse-mode \cite{franceschi2017forward}&$T$&$\mathcal{O}((P+H)T)$&$\frac{\partial \mathcal{L}_2}{\partial \alpha}+\frac{\partial \mathcal{L}_2}{\partial w_T}(\sum_{t=0}^{T}B_{t}A_{t+1}...A_T)$\\
Truncated Reverse-mode \cite{shaban2019truncated}&$K$&$\mathcal{O}((P+H)K)$&$\frac{\partial \mathcal{L}_2}{\partial \alpha}+\frac{\partial \mathcal{L}_2}{\partial w_T}(\sum_{t=T-K}^{T}B_{t}A_{t+1}...A_T)$\\
Neumann Series \cite{bengio2000gradient} &$\infty$&$\mathcal{O}(P+H)$&$\frac{\partial \mathcal{L}_2}{\partial \alpha}-\gamma \frac{\partial \mathcal{L}_2}{\partial w}\sum_{j=0}^{\infty}\left [ I- \gamma \frac{\partial^2 \mathcal{L}_1}{\partial w\partial w} \right ]^j \frac{\partial^2 \mathcal{L}_1}{\partial \alpha \partial w}$\\
Conjugate Gradient \cite{rajeswaran2019meta} &$S$&$\mathcal{O}(P+H)$&$\frac{\partial \mathcal{L}_2}{\partial \alpha}-\left ( \textup{argmin}_{\textup{x}}\left \| \textup{x} \frac{\partial^2 \mathcal{L}_1}{\partial w\partial w}-\frac{\partial \mathcal{L}_2}{\partial w}\right \| \right )\frac{\partial^2 \mathcal{L}_1}{\partial \alpha \partial w}$\\
Our Neumann approximation $\nabla_{\alpha}\hat{\mathcal{L}}_2^i$  &$K$&$\mathcal{O}(P+H)$&$\frac{\partial \mathcal{L}_2}{\partial \alpha}-\gamma \frac{\partial \mathcal{L}_2}{\partial w}\sum_{j=0}^{K}\left [ I- \gamma \frac{\partial^2 \mathcal{L}_1}{\partial w\partial w} \right ]^j \frac{\partial^2 \mathcal{L}_1}{\partial \alpha \partial w}$\\
\hline
\end{tabular}

\label{tab:hg_comp}
\end{table*}

\section*{E. Comparison of methods to approximate the hypergradient}

We compare different hypergradient approximations in Table \ref{tab:hg_comp}, which summarizes the computational complexity and memory cost for each method. Under the assumption that Hessian vector products are computed with the \textit{autograd}, we know that the compute time and memory cost for computing a
Hessian vector product are with a constant factor of the compute time and memory used for computing a single derivative $\frac{\partial \mathcal{L}_2}{\partial w}$ \cite{rajeswaran2019meta,griewank1993some,griewank2008evaluating}. We denote that the memory cost for computing the gradient of supernet weight $w$ and architecture parameters $\alpha$ are $P$ and $H$, respectively. We consider each step in the \textbf{Steps} means the computational time of computing a Hessian vector product. The Conjugate Gradient considers iterative solver (e.g., CG) to calculate the inverse of Hessian, where $S$ is the CG solver optimization steps, and each step contains the computation of Hessian vector product.

}

\clearpage
\end{document}